\documentclass[journal]{IEEEtran}
\usepackage{amsmath,amssymb,bm}
\usepackage{array,threeparttable,booktabs,multirow}
\usepackage{graphicx}
\usepackage{caption}
\usepackage{subcaption} 
\usepackage[ruled]{algorithm2e}
\usepackage[noend]{algpseudocode}
\usepackage{cite}
\usepackage[dvipsnames]{xcolor}
\usepackage{setspace}
\usepackage[colorlinks=true, allcolors=blue]{hyperref}
\usepackage{pifont}
\usepackage{makecell}
\usepackage{diagbox}
\usepackage{setspace}
\usepackage{float}   
\usepackage{enumitem}
\usepackage{orcidlink}
\usepackage{booktabs}         
\usepackage{siunitx}          
\usepackage{tikz}

\usepackage{pifont}

\newcommand{\revised}[1]{{\color{blue}#1}}
\newcommand{\cmark}{\checkmark}
\newcommand{\na}{\textemdash}   

\usepackage{booktabs}
\usepackage{multirow}
\usepackage{tabularx}
\usepackage{ragged2e} 
\usepackage{pifont}   
\usepackage{amssymb}  

\begin{document}

\title{

\huge EMFusion: Uncertainty-Aware Conditional Diffusion Model for Multivariate Narrow-band Exposure Forecasting
}
\author{
    Zijiang~Yan$^{*}$\orcidlink{0000-0002-7959-8329},
	Yixiang~Huang$^{*}$\orcidlink{0009-0000-4592-4491},~\IEEEmembership{Graduate~Student~Member,~IEEE}, %
	Jianhua~Pei\orcidlink{0000-0002-4066-9230}, ~\IEEEmembership{Member,~IEEE}, %
    Hina~Tabassum\orcidlink{0000-0002-7379-6949},~\IEEEmembership{Senior~Member,~IEEE,}
    and~Luca~Chiaraviglio\orcidlink{0000-0002-5350-2691},~\IEEEmembership{Senior~Member,~IEEE}

\thanks{
        Z. Yan and H. Tabassum are with the department of Electrical Engineering and Computer Science, York University, Toronto, Canada (e-mails:  
        \{\href{mailto:zijiang@yorku.ca}{zijiang}, \href{mailto:hinat@yorku.ca}{hinat}\}@yorku.ca). 
		Y. Huang and J. Pei are with the School of Electrical and Electronic Engineering, Huazhong University of Science and Technology, Wuhan, China (e-mails: \href{mailto:yxhuang@hust.edu.cn}{yxhuang@hust.edu.cn}, \href{mailto:jianhuapei98@gmail.com}{jianhuapei98@gmail.com}).
        Y. Huang is also with China Southern Grid Electric Power Dispatch Control Center, Guangzhou, China.
        J. Pei is also with the Central China Branch of State Grid Corporation of China, Wuhan, China. 
        L. Chiaraviglio is with the Department of Electronic Engineering, University of Rome Tor Vergata, Rome, Italy (e-mail: \href{mailto:luca.chiaraviglio@uniroma2.it}{luca.chiaraviglio@uniroma2.it}). 
        L. Chiaraviglio is also with the Consorzio Nazionale Interuniversitario per le Telecomunicazioni (CNIT), Parma, Italy.
        \*Z. Yan and Y. Huang contributed equally to this work and are co-first authors.

        This work was supported in part by Natural Sciences and Engineering Research Council of Canada (NSERC) Discovery grant and in part by European Research Council (ERC) under European Union’s Horizon Europe Grant 101078411.
        
	}%

}


\raggedbottom

\maketitle

\begin{abstract}
The rapid growth in wireless infrastructure has increased the need to accurately estimate and forecast electromagnetic field (EMF) levels to ensure ongoing compliance, assess potential health impacts, and support efficient network planning.  While existing studies rely on univariate forecasting of wideband aggregate EMF data, multivariate narrow-band EMF forecasting is needed to capture the inter-operator and inter-frequency variations essential for proactive network planning. To this end, this paper introduces EMFusion, a conditional diffusion-based EMF forecasting framework that integrates diverse contextual factors (e.g., time of day, season, and holidays) while providing uncertainty-aware probabilistic forecasts. The proposed architecture features a residual U-Net backbone enhanced by a {cross-attention mechanism} that dynamically integrates external conditions to guide the generation process. Furthermore,
EMFusion integrates an {imputation-based sampling strategy} that treats forecasting as a structural inpainting task, ensuring temporal coherence even with irregular measurements. Unlike standard point forecasters, 
EMFusion generates empirical probabilistic prediction intervals from the learned conditional distribution, providing uncertainty-aware probabilistic forecasting rather than simple point estimation.
Numerical experiments conducted on the multivariate narrow-band  EMF datasets demonstrate that EMFusion with the contextual information of working hours outperforms the baseline models with or without conditions.  {The proposed EMFusion outperforms the best baseline by  23.85\% in continuous ranked probability score (CRPS) and 13.93\% in normalized root mean square error.}
\end{abstract}

\section{Introduction} 
\label{sec:introduction}
The rapid proliferation of new wireless communication technologies, such as transmissions at higher frequencies, massive antennas, and reconfigurable intelligent surfaces (RISs), has led to an unprecedented increase in the radio-frequency (RF)-based electromagnetic field (EMF) sources \cite{sadeq, chiaraviglio2021health}. 
This technological surge has amplified public concerns regarding potential human health effects associated with EMF exposure \cite{chiaraviglio2021health, icnirp2020guidelines,chiaraviglio2022much}. Consequently, accurate EMF monitoring and forecasting are becoming indispensable for regulatory compliance, network planning, and resource management \cite{mootoo2025emforecaster, stam2018comparison,pei2025latent}.

\subsection{Motivation}

Despite the advances in EMF modeling and monitoring, EMF forecasting efforts remain limited \cite{sambo2016electromagnetic, jiang2023rate,ibraiwish2021emf,chiaraviglio2018planning, ITU:2019}. The handful of EMF forecasting studies remain primarily focused on univariate prediction of aggregate (wideband) exposure across large bandwidths \cite{mootoo2025emforecaster,kiouvrekis2024comparative, pala2021examining, nguyen2024deep,al2023machine}. However, wideband exposure alone is often insufficient for regulatory oversight, active network planning, or proactive resource allocation. In practice, a narrow-band EMF breakdown is required to identify dominant exposure sources and quantify the contributions of specific technologies (e.g., 3G, 4G, and 5G deployments) and network operators. From a regulatory perspective, this granularity is essential. Environmental protection agencies must enforce maximum exposure limits at each frequency band across the territory to safeguard public health\footnote{Simulation-based assessments submitted during site authorization are typically conservative and may differ from actual measured exposure levels, as operators often declare maximum radiated power, antenna tilting, or beamforming configurations that exceed those used in operation \cite{chiaraviglio2022much}.}. Extending monitoring capabilities to multivariate narrow-band forecasting would enable forward-looking compliance verification and transform periodic reporting from retrospective documentation into proactive decision support. 

Moreover, per-frequency forecasts could establish a feedback loop between regulatory oversight and network management, allowing operators to adjust spectrum allocation, transmission power, or deployment strategies, particularly in sensitive areas, before exposure limits are approached.
{More specifically, forecasting outputs can be integrated into network optimization and control pipelines by treating predicted EMF exposure
$\widehat{\mathbf{X}}_{t+h}(\boldsymbol{\theta})$ as a future operational constraint, i.e.,
$\max_{\boldsymbol{\theta}} \: U(\boldsymbol{\theta}) \ \text{s.t.}\
\widehat{\mathbf{X}}_{t+h}(\boldsymbol{\theta}) \leq \mathbf{X}_{\max},\
\boldsymbol{\theta} \in \mathcal{C}$,
where $\boldsymbol{\theta}$ denotes network control variables (e.g., transmit power, beamforming, and scheduling), $U(\boldsymbol{\theta})$ represents a network utility function, $\mathbf{X}_{\max}$ is the regulatory EMF threshold, and $\mathcal{C}$ captures standard QoS and power constraints.} In this way, operators can anticipate when and in which frequency bands EMF exposure may approach regulatory limits, enabling threshold-based alerts, adaptive scheduling, and exposure-aware resource management before compliance violations occur.
While these examples illustrate applications of EMF forecasting, this work focuses  on forecasting and uncertainty quantification; closed-loop network-control validation is left for future work.

\subsection{Research Challenges}
Multivariate narrow-band EMF forecasting introduces several challenges. First, EMF measurements across different frequencies, operators, and wireless technologies exhibit correlated temporal dynamics that cannot be adequately captured by aggregate wideband forecasting. Second, practical network planning requires uncertainty-aware predictions rather than deterministic point estimates. Third, EMF exposure is influenced by contextual factors such as working hours, weekdays, holidays, and seasonal patterns. Finally, real-world monitoring systems often contain missing or irregular measurements due to sensor outages and operational constraints.

Although deep learning (DL) has significantly advanced time-series forecasting, conventional architectures such as recurrent neural networks (RNNs) and transformers are primarily designed for deterministic prediction. Consequently, they often produce overconfident forecasts, struggle to quantify uncertainty, and may accumulate errors over long prediction horizons. In contrast, generative diffusion models (DMs) learn the full distribution of future trajectories through an iterative denoising process \cite{huang2025probabilistic, sohl2015deep, ho2020denoising}. Moreover, DMs naturally support data imputation \cite{ref:imputation, 10896580} and flexible conditioning on contextual information \cite{borovykh2017conditional}, making them well-suited for modeling the uncertainty of  EMF dynamics. Building on the need for narrow-band and uncertainty-aware EMF forecasting, this paper proposes \textit{EMFusion}, a conditional diffusion model (CDM) for multivariate probabilistic forecasting of EMF exposure across multiple frequencies. Since narrow-band EMF measurements are performed across multiple frequency bands, wireless technologies, and network operators, the resulting data are inherently multivariate and exhibit correlated exposure dynamics. EMFusion models the joint distribution of these multi-frequency trajectories while providing uncertainty quantification directly through its generative process.

\begin{table*}[t]
\centering
\small 
\caption{Classification of deep learning models for wireless networks forecasting/prediction applications. \textbf{UV}: Univariate forecasting; \textbf{MV}: Multivariate forecasting; \textbf{Cond.}: External Conditions; \textbf{Imp.}: Imputation; \textbf{Trust.}: Trustworthiness.}
\label{tab:wireless_refs_ungrouped}

\setlength{\tabcolsep}{0.3em} 
\renewcommand{\arraystretch}{1.2}

\begin{tabularx}{\linewidth}{l l >{\RaggedRight}p{4cm} >{\RaggedRight\arraybackslash}X c c c c}
\toprule
\textbf{Domain} & \textbf{Ref.} & \textbf{Model} & \textbf{Target Dataset} & \textbf{UV/MV} & \textbf{Cond.} & \textbf{Imp.} & \textbf{Trust.} \\
\midrule

\multirow{6}{*}{\shortstack[l]{\textbf{Network}\\\textbf{Traffic}}}
& \cite{nikravesh2016mobile}
& MLP, MLPWD, SVM 
& Commercial LTE trial traffic 
& UV/MV 
& \na
& \na 
& \na \\

& \cite{di2023multivariate} 
& VAR, RNN 
& LTE-A VoIP traffic 
& UV/MV 
& \na
& \na 
& \na \\

& \cite{dalgkitsis2018traffic} 
& LSTM RNN 
& Vodafone 4G throughput 
& UV 
& \na
& \na 
& \na \\

& \cite{habib2024transformer} 
& Transformer
& 5G/LTE O-RAN aggregate traffic 
& UV 
& \na
& \na 
& \na \\

& \cite{fiandrino2024aichronolens} 
& LSTM + XAI 
& 4G load and active-user datasets 
& UV 
& \na
& \na 
& \na \\

& \cite{isravel2024multivariate} 
& VAR, LSTM 
& ubiquitous healthcare traffic 
& MV 
& \na
& \na 
& \na \\
\midrule

\multirow{2}{*}{\shortstack[l]{\textbf{Channel}\\\textbf{Prediction}}}
& \cite{sone2020wireless} 
& Holt--Winters, LSTM 
& Real enterprise WLAN traffic 
& UV/MV 
& BS Traffic
& \na 
& \na \\

& \cite{zhang2021deep} 
& LSTM (SCP), RNN 
& Synthetic LEO mMIMO CSI 
& UV 
& \na
& \na 
& \na \\
\midrule

\multirow{6}{*}{\shortstack[l]{\textbf{QoS}\\\textbf{Prediction}}}
& \cite{hameed2022toward} 
& Temporal Transformer 
& Real IoT smart-building datasets 
& UV/MV 
& \na
& \textbf{\cmark} 
& \na \\

& \cite{dinaki2021forecasting} 
& BiLSTM--CNN 
& CDN video streaming QoE dataset 
& MV 
& \na
& \na 
& \na \\

& \cite{colpitts2023short} 
& LSTM, Seq2Seq 
& Rural fixed-wireless LTE KPIs 
& MV 
& Weather
& \textbf{\cmark} 
& \na \\

& \cite{zhang2025probabilistic} 
& Diffusion + TCN 
& Real DTN-inspired sensor datasets 
& MV 
& Tempreature
& \na 
& \textbf{\cmark} \\
\midrule

\multirow{7}{*}{\shortstack[l]{\textbf{EMF}\\\textbf{Exposure}}}

& \cite{song2025study} 
& LSTM, CNN
& EMF measurements 
& UV 
& Population Density
& \na 
& \na \\

& \cite{kiouvrekis2024comparative} 
& k-NN, XGBoost 
& EMF measurements 
& UV 
& Population Density
& \na 
& XAI \\

& \cite{bakcan2022measurement} 
& CNN
& Wideband  RF-EMF 
& UV 
& \na
& \na 
& \na \\

& \cite{pala2021examining} 
& LSTM, RNN
& Wideband  RF-EMF 
& UV 
& \na
& \na 
& \na \\

& \cite{nguyen2024deep} 
& Transformer, CNN 
& Wideband  RF-EMF 
& MV 
& \na
& \na 
& \na \\

& \cite{mootoo2025emforecaster} 
& Patching, MLP 
& Wideband RF-EMF 
& UV 
& \na
& \na 
& \cmark \\

\addlinespace 
& \textbf{EMFusion} 
& \textbf{CDM + Cross attention}
& Narrow-band Multivariate RF-EMF
& \textbf{UV/MV} 
& \textbf{Time, Date, Season}
& \textbf{\cmark} 
& \textbf{\cmark} \\

\bottomrule
\end{tabularx}
\end{table*}


\subsection{Contributions}
To the best of our knowledge, conditional diffusion models have not been applied to time-series forecasting in wireless networks in general, nor to EMF forecasting in particular. 
The specific contributions of this paper can thus be summarized as:
\begin{enumerate}
    \item  The proposed \textit{EMFusion} is based on a CDM  for multivariate probabilistic forecasting of frequency-specific EMF exposure that allows the generation process to be guided by various contextual factors or conditions.  \textit{EMFusion} leverages contextual variables such as time of the day, day of the week, holiday/working days, working hours, and seasonal trends. This conditioning not only enhances the accuracy and relevance of the forecasts, but also allows for the generation of scenarios tailored to specific environmental states. 
    The proposed architecture features a residual U-Net backbone enhanced by a {cross-attention mechanism} that dynamically integrates external conditions to guide the generation process. Furthermore, EMFusion integrates an {imputation-based sampling strategy} that treats forecasting as a structural inpainting task, ensuring temporal coherence even with irregular or missing EMF measurements. 
    

    \item {
    Unlike conventional point forecasters that output a single trajectory, we propose a probabilistic interval estimation framework based on the generative capabilities of EMFusion. By treating the ensemble of generated trajectories as samples from a conditional distribution, we employ Kernel Density Estimation (KDE) to reconstruct continuous probability density functions at each time step. This approach enables the derivation of empirical prediction intervals without imposing parametric Gaussian assumptions, thus achieving improved uncertainty quantification.
    }
    
    \item \textit{EMFusion} can be customized for both univariate and multivariate forecasting scenarios. Multivariate forecasting captures correlation across distinct network operators, transmission frequencies, and wireless technologies. Meanwhile, a univariate approach remains beneficial for scenarios where data or computational resources are limited, enabling fast, frequency-specific forecasts that are simpler to implement for localized optimization.
     \item We conduct experiments on multivariate narrow-band EMF exposure datasets spanning multiple narrow-band channels in the frequency range of 9 kHz–6 GHz, thus covering major network operators in the Italy and cellular network technologies.  Numerical results confirm the effectiveness of the conditional over unconditional EMFusion, and multivariate over univariate EMFusion. Among various exogenous conditions, we note that the working hour condition outperforms. {Furthermore, the EMFusion outperforms the best baseline by 23.85\% in continuous ranked probability 
score (CRPS) and 13.93\% in normalized root mean square error (NRMSE).}
\end{enumerate}

The remainder of this paper is organized as follows.
Section~\ref{sec:2} presents the existing state-of-the-art for TSF in wireless networks. Section~\ref{sec:uemff} presents the architecture of the proposed multivariate EMFusion model. The experimental setup, data analysis, and visualizations are presented in Section~\ref{sec:expierment_data_visualization} followed by numerical results and discussions in Section~\ref{sec:results}. Finally, Section \ref{sec:Conclusion} concludes the paper.

\section{Related Work}
\label{sec:2}
This section reviews the literature most relevant to multivariate narrow-band EMF forecasting. We first discuss EMF modeling, monitoring, and exposure-assessment frameworks, followed by time-series forecasting methods in wireless networks, including statistical, deep learning, and generative AI approaches. The review highlights the methodological foundations that motivate the proposed EMFusion framework.


\subsection{EMF Modeling and Monitoring}

 Existing studies primarily rely on stochastic modeling and optimization frameworks to manage exposure. Representative approaches include resource allocation and energy-efficiency maximization under EMF limits \cite{sambo2016electromagnetic, jiang2023rate}, the integration of RIS to reduce exposure while maintaining service quality \cite{ibraiwish2021emf}, and EMF-aware network planning through optimized base station (BS) placement and configuration \cite{chiaraviglio2018planning, ITU:2019}.
Operational strategies, such as exposure-conscious user association \cite{matalatala2018joint}, beamforming optimization \cite{ying2013beamformer}, and adaptive cross-layer protocols \cite{nadas2019reducing}, further aim at exposure mitigation.

Beyond modeling-based strategies, some recent works have strengthened EMF measurement methodologies and long-term monitoring in the context of 5G deployment. For instance, \cite{franci2020experimental} and \cite{migliore2022application} developed rigorous procedures for estimating the instantaneous maximum received power from 5G sources and refining exposure assessment at millimeter-wave frequencies. The IEC 62232:2025 standard \cite{IEC62232_2025} reports case studies evaluating EMF exposure across typical BS scenarios from 100 kHz to 300 GHz. Longitudinal monitoring data in \cite{iakovidis20255g} reveal a gradual increase in exposure at 3.6 GHz over two years in Greece, linked to expanding 5G infrastructure.

\subsection{Time-Series Forecasting in Wireless Networks}
\subsubsection{Statistical Methods for TSF}

Initial approaches to EMF exposure forecasting relied on classical time-series models. These included autoregressive models like ARIMA~\cite{box2015time}, as well as methods such as kernel ridge regression and support-vector regression. While computationally lightweight and interpretable, these techniques often assume stationarity and require significant manual feature engineering. Their linear parametrisation also limits their accuracy when the underlying drivers of EMF exposure interact non-linearly across multiple time scales.
A representative approach is found in \cite{de2019time}, who proposed a model that decomposes wide-band EMF radiation into three distinct components: the average exposure level, daily and half-daily periodic patterns (seasonal behavior), and the temporal correlation among residuals \cite{de2019time}. Their model effectively captured different behaviors between weekdays and weekends by applying separate autoregressive model for weekdays and Sunday, and the other autoregressive model for Saturday \cite{de2019time}. While effective, such methods are tailored to specific observed periodicities and struggle to generalize without manual recalibration.

\subsubsection{Deep Learning for Time-Series Forecasting}
{
\paragraph{Network Traffic Forecasting}
\cite{nikravesh2016mobile} compared Multi-Layer Perceptrons (MLP)  as well as Support Vector Machines (SVM), to predict traffic in commercial Long-Term Evolution (LTE) networks. Similarly, \cite{dalgkitsis2018traffic} developed a univariate forecasting platform for 4G networks using LSTM and Recurrent Neural Networks (RNNs), demonstrating superiority over statistical baselines like ARIMA. More recent studies have shifted toward multivariate and interpretable models. \cite{di2023multivariate} characterized LTE-Advanced Voice-over-IP metrics using Vector Autoregression (VAR) and supervised DL sliding-window frameworks. To enhance interpretability, \cite{fiandrino2024aichronolens} introduced AICHRONOLENS, an Explainable AI (XAI) framework that correlates temporal structures with LSTM error diagnoses. In the context of Software-Defined Networking, \cite{isravel2024multivariate} utilized LSTM to model jointly evolving traffic flows for healthcare applications. Furthermore, for open radio access networks, \cite{habib2024transformer} proposed a Transformer-based Autoformer architecture to predict aggregate cell traffic.
While these methods achieve strong deterministic accuracy, they primarily focus on point forecasting. They generally lack explicit mechanisms for data imputation handling missing values and do not provide intrinsic uncertainty quantification essential for uncertainty-aware network planning.

\paragraph{Channel Prediction}
Channel prediction is critical for mitigating aging in high-mobility scenarios. \cite{sone2020wireless} benchmarked various models, including Holt-Winters and XGBoost, against LSTMs for enterprise Wireless Local Area Network traffic, analyzing spatio-temporal Access Point (AP) clustering. In satellite communications, \cite{zhang2021deep} proposed a Satellite Channel Predictor based on LSTM units for Low Earth Orbit (LEO) massive Multiple-Input Multiple-Output (mMIMO) systems. The approach outperformed Kalman filters and standard RNNs in multi-step Channel State Information (CSI) forecasting.

\paragraph{Quality of Service (QoS) Prediction}
Forecasting QoS involves modeling complex dependencies. \cite{dinaki2021forecasting} introduced a hybrid bidirectional LSTM (BiLSTM) and Convolutional neural network (CNN) architecture for multivariate QoS forecasting in Content Delivery Networks (CDNs). \cite{hameed2022toward} utilized a temporal Transformer encoder to predict QoS metrics in IoT networks, effectively capturing long-term dependencies that RNNs often miss. \cite{colpitts2023short} investigated Sequence-to-Sequence (Seq2Seq) models for rural LTE Key Performance Indicators (KPIs), noting that exogenous features provided minimal gains due to short temporal dependencies.

\paragraph{EMF Exposure Forecasting}
Research in EMF forecasting has progressed from spatial regression to time-series analysis. \cite{kiouvrekis2024comparative} focused on comparative regression using ensemble methods like Random Forest  to rank urban factors affecting EMF levels. For temporal forecasting, \cite{pala2021examining} demonstrated that LSTMs outperform classical statistical models (e.g., ARIMA) in long-term (over a 60-month period) for univariate EMF prediction. Recently, \cite{nguyen2024deep} demonstrated the effectiveness of 1-D CNNs in handling single-step and multi-step EMF forecasting tasks by leveraging translation invariance to identify periodicities in the data. Other research works also adapted CNN designs, utilizing architectures like Radial Basis Function Networks and Generalized Regression Neural Networks to model complex RF-EMF signals from emerging technologies like  5G~\cite{al2023machine}. Despite these advances, existing EMF forecasting studies remain largely focused on deterministic or univariate prediction of aggregate exposure levels. Consequently, challenges related to narrow-band multivariate forecasting, uncertainty quantification, and robustness to incomplete observations remain insufficiently explored.

These limitations have motivated growing interest in probabilistic generative models capable of learning complex temporal distributions and providing uncertainty-aware forecasts.
}

\subsubsection{Generative AI for TSF}
 Generative models have emerged as a powerful paradigm within DL for TSF. Representative architectures include Variational Autoencoders (VAEs)~\cite{li2022generative}, Generative Adversarial Networks (GANs)~\cite{goodfellow2014generative,  koochali2021if}, and, more recently, DMs ~\cite{song2019generative, ho2020denoising}, which have gained remarkable attention for their ability to capture data uncertainty and generate high-fidelity forecasts. DMs, as latent-variable generative frameworks, learn complex data distributions via a dual process of progressive noise injection (the \textit{forward diffusion}) and iterative denoising (the \textit{reverse process}). 
More recently, DMs have been successfully adapted to time series data~\cite{rasul2021autoregressive, tashiro2021csdi}, offering a flexible generative framework for learning intricate temporal dependencies and stochastic dynamics.

%
While DiffTCN introduces probabilistic modeling, the majority of QoS research remains deterministic. 


Compared to conventional deterministic or probabilistic models, DMs present several advantages:
\begin{enumerate}
    \item DMs yield fully probabilistic forecasts, enabling uncertainty quantification through multiple stochastic samples and predicting extreme cases in terms of different EMF traffic flows, whereas deterministic models are disadvantaged in capturing uncertainty  \cite{yan2025semantic}.
    \item They maintain stable and well-behaved training dynamics, avoiding adversarial divergence;
    \item They offer a highly modular design that allows flexible conditioning on exogenous factors such as seasonal indicators, control variables, or multimodal inputs~\cite{kollovieh2023predict, yuan2024diffusion, liu2024retrieval}.
\end{enumerate}

Empirically, DMs demonstrate competitive or even state-of-the-art results across numerous TSF benchmarks~\cite{shen2024multi, liu2024retrieval, li2025diffusion,meijer2024rise,su2025diffusion}, positioning them as a cornerstone of next-generation generative forecasting frameworks.

\begin{figure*}[!ht]
\vspace{-0.1cm}
\centerline{\includegraphics[width=0.75\textwidth]{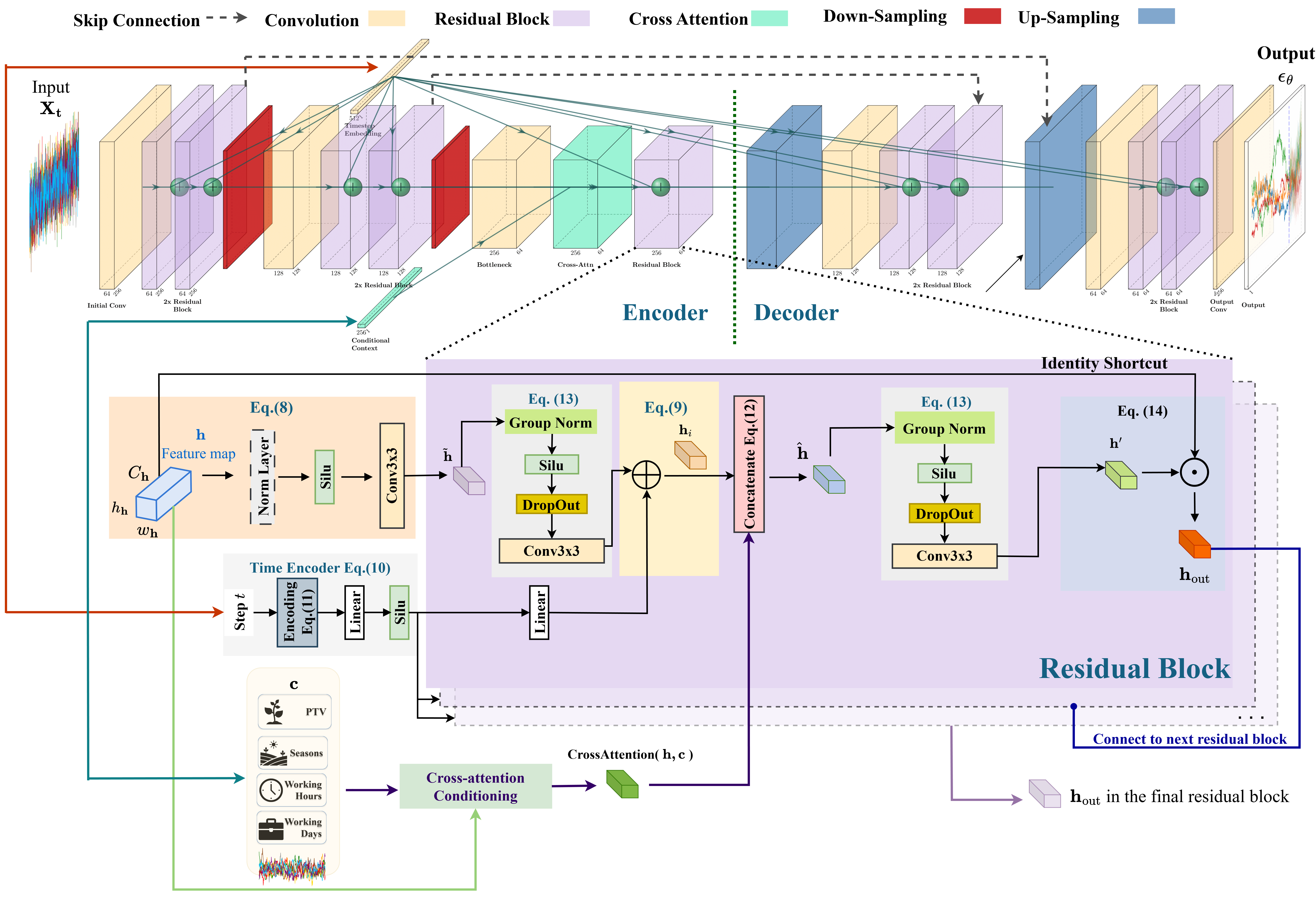}}
\caption{ The U-Net architecture used for EMFusion. The network processes a noisy input through a symmetric encoder-decoder path. Timestep embeddings are incorporated into the residual blocks, while external context is injected via a cross-attention block {between the residual blocks of the encoder and decoder}. The residual block architecture of EMFusion to estimate $\epsilon_\theta(\cdot)$. 
}
\label{fig:residual-block}
\vspace{-0.2cm}
\end{figure*}

\section{ EMFusion: Conditional DM for Multivariate Narrow-Band EMF Forecasting}
\label{sec:uemff}

This section details our methodology for multivariate EMF TSF leveraging a conditional DM. Given the sequence of historical EMF measurements and corresponding contextual factors,  the goal is to model the conditional distribution of the future EMF values. The proposed model is then enhanced with imputation-based resampling to deal with real-world EMF logs where  EMF measurements may be irregularly sampled or there are gaps or sensor outages. 
{
Before detailing the formal mathematical framework, we provide a conceptual overview of the three core pillars of the EMFusion architecture:

\begin{itemize}
    \item \textbf{Diffusion Modeling as Iterative Refinement:} Unlike standard models that predict a single value, diffusion modeling treats forecasting as a process of systematic denoising. The model learns to reconstruct a structured, high-fidelity EMF signal from Gaussian noise through multiple refinement steps \cite{ho2020denoising, huang2025probabilistic}.
    
    \item \textbf{Cross-Attention as Dynamic Weighting:} To ensure forecasts are context-aware, we utilize a cross-attention mechanism. This acts as a dynamic lens, allowing the model to selectively focus on specific historical patterns or external conditions (e.g., working hours or seasons) that are most relevant to the EMF time-series \cite{huang2025probabilistic, 10896580}.
    
    \item \textbf{Inpainting as Pattern Completion:} We frame the forecasting task as a structural inpainting problem. Much like filling in a missing piece of a photograph, the model treats future EMF values as the missing portion of a temporal image and utilizes the known "texture" of the past to complete the sequence realistically \cite{ref:imputation}.
\end{itemize}
}

\subsection{Multivariate Conditional Diffusion Model}
\label{sec:emf_conditional_generation}
\begin{figure*}[!ht]
\vspace{-0.1cm}
\centerline{\includegraphics[width=0.75\linewidth]{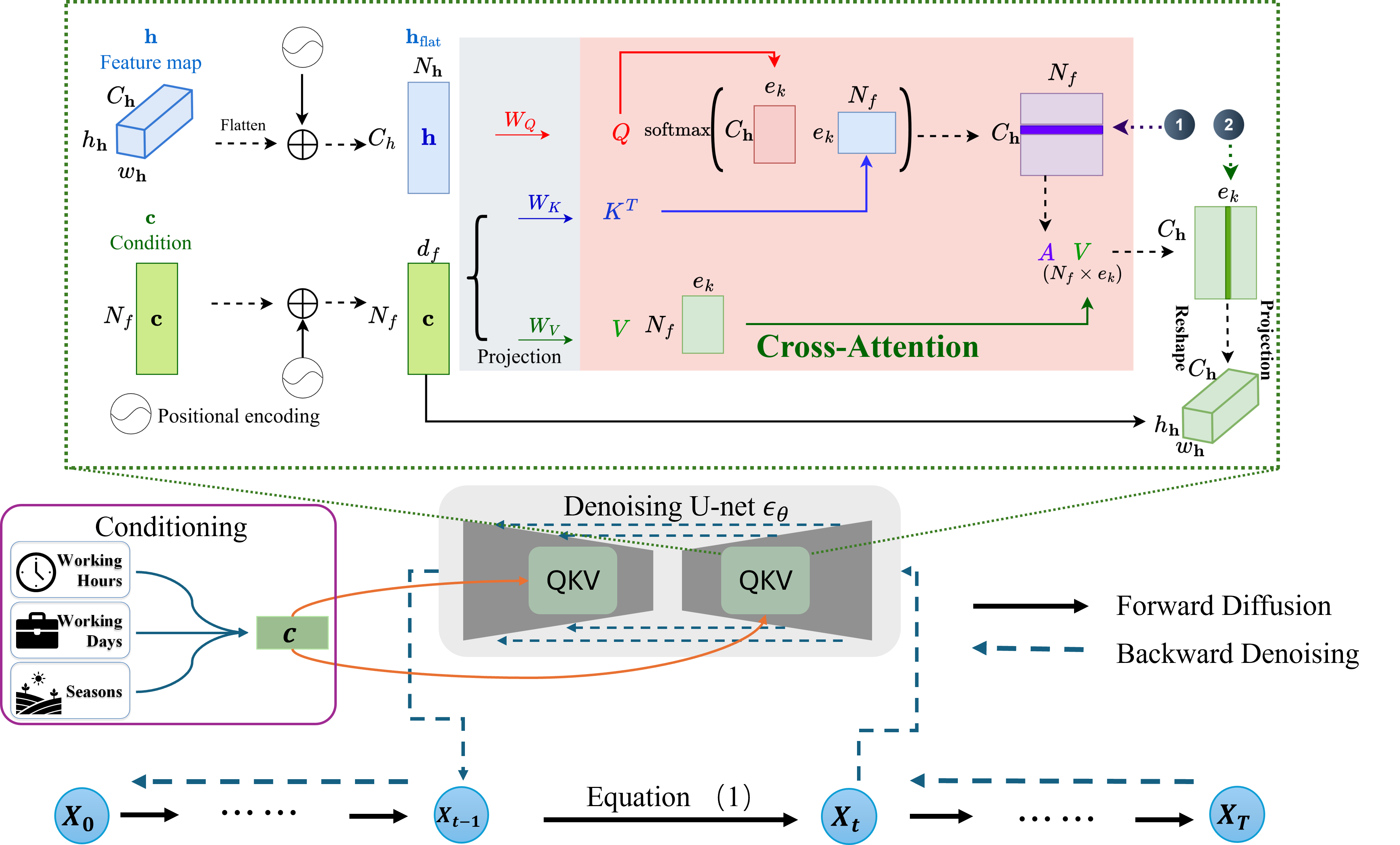}}
\caption{EMFusion for multivariate frequency selective EMF forecasting with cross-attention. }
\label{fig:uv-cross-attention}
\vspace{-0.2cm}
\end{figure*}
We consider a multivariate time-series forecasting problem in which each 
observation $\mathbf{x}_k \in \mathbb{R}^N$ represents measurements from $N$ variates or frequency channels at time~$k$. Given a historical window of length $H$, denoted by 
$\mathbf{X}_{\text{past}} = (\mathbf{x}_{-(H+1)}, \ldots, \mathbf{x}_0)$, 
The objective is to predict the future multichannel trajectory 
$\mathbf{X}_{\text{future}} = (\mathbf{x}_{1}, \ldots, \mathbf{x}_{F})$ over 
a forecast horizon~$F$. To model the conditional distribution 
$p_\theta(\mathbf{X}_{\text{future}} \mid \mathbf{X}_{\text{past}}, \mathbf{c})$, 
we introduce an external condition vector $\mathbf{c}$, which may include 
calendar features, operational metadata, or other exogenous covariates. Unlike 
vectorized formulations, we treat the full future window as a matrix 
{$\mathbf{X}_0 := [\mathbf{X}_{\text{past}}, \mathbf{X}_{\text{future}}] \in \mathbb{R}^{(H+F) \times N}$ }and 
apply diffusion directly in this multivariate tensor space. This enables 
probabilistic forecasting while preserving the temporal–channel structure of the 
signal.

In the diffusion-based formulation, the forward noising process corrupts the 
clean  $\mathbf{X}_0$ over $T$ time steps according to the Markov transition process given by \cite{ho2020ddpm, huang2025probabilistic}:
\begin{equation}
    q(\mathbf{X}_t \mid \mathbf{X}_{t-1})
    = \mathcal{N}\!\left(\sqrt{\alpha_t}\,\mathbf{X}_{t-1},
    (1-\alpha_t)\mathbf{I}\right),
\end{equation}
which admits the closed-form marginal
\begin{equation}
    q(\mathbf{X}_t \mid \mathbf{X}_0)
    = \mathcal{N}\!\left(\sqrt{\bar{\alpha}_t}\,\mathbf{X}_0,\,
    (1-\bar{\alpha}_t)\mathbf{I}\right),
\end{equation}
where $\alpha_t \in (0,1]$ is a predefined noise schedule, $\mathbf{I}$ denotes the identity matrix, and $\bar{\alpha}_t = \prod_{s=1}^{t} \alpha_s$ {, where $\alpha_s$ is the noise intensity of each step in the noise addition process}. During training, noisy future 
samples can be directly sampled from $\mathbf{X}_0$ at any timestep $t$, i.e.,
\begin{equation} \label{eq:x_t}
    \mathbf{X}_t 
    = \sqrt{\bar{\alpha}_t}\,\mathbf{X}_0
      + \sqrt{1-\bar{\alpha}_t}\,\boldsymbol{\epsilon},
    \qquad
    \boldsymbol{\epsilon} \sim \mathcal{N}(0,\mathbf{I}),
\end{equation}
with $\boldsymbol{\epsilon} \in \mathbb{R}^{(H+F) \times N}$ containing i.i.d.\ 
Gaussian noise.

The training objective minimizes the discrepancy between the true noise $\boldsymbol{\epsilon}$ and the network-predicted noise $\boldsymbol{\epsilon}_\theta(\mathbf{X}_t, t, \mathbf{c})$ using a mean-squared error loss:
\begin{equation}
\label{eq:loss_training}
L(\theta) = \mathbb{E}_{t, \mathbf{X}_0, \boldsymbol{\epsilon}}
\!\left[
\big\|
\boldsymbol{\epsilon} -
\boldsymbol{\epsilon}_\theta\!\left(
\sqrt{\bar{\alpha}_t}\,\mathbf{X}_0 +
\sqrt{1 - \bar{\alpha}_t}\,\boldsymbol{\epsilon},
t, \mathbf{c}
\right)
\big\|^2
\right].
\end{equation}

The reverse diffusion process serves as the generative mechanism for forecasting. The conditional reverse diffusion process predicts the distribution
{$p_\theta(\mathbf{X}_{t-1} \mid \mathbf{X}_t, \mathbf{c})$}, of the less noisy state $\mathbf{X}_{t-1}$ given the current noisy state $\mathbf{X}_t$ and the conditioning information $\boldsymbol{c}$. It learns to reconstruct the original signal $\mathbf{X}_0$ by iteratively denoising the noisy samples, starting from pure noise $\mathbf{X}_T \sim \mathcal{N}(0, \mathbf{I})$ \cite{pei2026deep}.  A neural network $\boldsymbol{\epsilon}_{\theta}(\mathbf{X}_t, t, \boldsymbol{c})$ is employed to approximate the reverse transition distribution $p_\theta(\mathbf{X}_{t-1} | \mathbf{X}_t, \mathbf{c})$ by predicting the noise component $\boldsymbol{\epsilon}$ that was added at step $t$. This noise prediction network takes the noisy data $\mathbf{X}_t$, the current timestep $t$, and the conditioning vector $\mathbf{c}$ to predict the following conditional distribution:
\begin{equation} \label{2}
    p_\theta(\mathbf{X}_{t-1} \mid \mathbf{X}_t,
              \mathbf{c})
    = \mathcal{N}\!\bigl(
        \boldsymbol{\mu}_\theta(\mathbf{X}_t, t,
                                 \mathbf{c}),
        \boldsymbol{\Sigma}_t (\mathbf{X}_t, t,
                                 \mathbf{c})
      \bigr),
\end{equation}
where the mean  computed from 
$\boldsymbol{\epsilon}_\theta$ is given by:
\begin{equation}
    \boldsymbol{\mu}_\theta(\mathbf{X}_t, t,
                            \mathbf{c})
    = \frac{1}{\sqrt{\alpha_t}}
      \left(
        \mathbf{X}_t 
        - \frac{1-\alpha_t}{\sqrt{1-\bar{\alpha}_t}}\,
          \boldsymbol{\epsilon}_\theta(\mathbf{X}_t, t, \mathbf{c})
      \right).
\end{equation}
and $\mathbf{c}$ encapsulates features from the dataset, such as \texttt{Italy\_WorkingDay}, \texttt{Italy\_WorkingHour}, and \texttt{Italy\_Season}, which depict factors influencing EMF variations \cite{meijer2024rise,su2025diffusion}. By conditioning the noise prediction on  $\mathbf{c}$, the model learns to generate EMF trajectories consistent with temporal continuity and relevant external dependencies. 

This formulation enables a fully multivariate, conditionally guided reverse 
diffusion process in which $\mathbf{X}_t$ is iteratively denoised using both the 
historical multichannel window $\mathbf{X}_{\text{past}}$ and the external 
condition vector $\mathbf{c}$.

{
During inference, a classifier-free guidance (CFG) mechanism \cite{ref:classifier-free,huang2025probabilistic} steers the generation process. The final noise prediction is obtained by combining the conditional and unconditional outputs of the denoising network:
\begin{equation}
\hat{\boldsymbol{\epsilon}}_\theta(\mathbf{X}_t, t, \mathbf{c}) =
(1 + s_c)\,\boldsymbol{\epsilon}_\theta(\mathbf{X}_t, t, \mathbf{c})
- s_c\,\boldsymbol{\epsilon}_\theta(\mathbf{X}_t, t, \emptyset),
\end{equation}
where $s_c$ is the guidance scale controlling the adherence to the conditioning input $\boldsymbol{c}$, and $\emptyset$ represents an unconditional prediction. In conditional TSF, positive CFG scales often over-concentrate the posterior, which narrows prediction intervals. Since our model already injects conditioning via cross-attention, we found {$s_c =0 $} is the best via hyperparameter tuning. 
This conditional diffusion formulation provides the theoretical basis for our model. 
}

\begin{algorithm}[t]
\caption{DDPM Training for Multivariate Forecasting with Cross-Attention}
\KwIn{ {$\mathbf{X}_{0}=[\mathbf{X}_{\text{past}}, \mathbf{X}_{\text{future}}]$, conditions $\mathbf{c}$}}
\KwOut{Learned parameters $\theta$}


Sample $t \sim \mathcal{U}\{1,\ldots,T\}$ and noise $\boldsymbol{\epsilon} \sim \mathcal{N}(0,I)$\;
Generate noisy sequence: $\mathbf{X}_t=\sqrt{\bar{\alpha}_t}\mathbf{X}_{0}+\sqrt{1-\bar{\alpha}_t}\boldsymbol{\epsilon}$\;
Pass $\mathbf{X}_t$ and $t$ through U-Net\;
{ Time series \revised{$t$} embedding}\\
\For{each cross-attention block}{
{
{Positional encoding computed by Eq (\ref{eq:positional_encoding})}\;}
Compute $Q=\mathbf{h}_{\mathrm{flat}}W_Q$, $K=\mathbf{c}W_K$, $V=\mathbf{c}W_V$\;
Compute $\text{Attention}=\mathrm{Softmax}(QK^\top/\sqrt{e_k})V$\;
Update feature map: $\mathbf{h}\leftarrow\mathbf{h}+\text{Attn}$\;
}
Predict noise {with Cross Attention by Eq~(\ref{eq:residual_addition})}: $\hat{\boldsymbol{\epsilon}}_\theta=\epsilon_\theta(\mathbf{X}_t,t,\mathbf{c})$\;
Update parameters by minimizing $\mathcal{L}(\theta)=\|\boldsymbol{\epsilon}-\hat{\boldsymbol{\epsilon}}_\theta\|_2^2$\;
\label{alg:training}
\end{algorithm}

\begin{algorithm}[t]
\caption{DDPM Inference for Multivariate Forecasting with Cross-Attention and Masked Imputation}
\KwIn{ $\mathbf{X}_{\mathrm{past}}$, mask $\boldsymbol{\Omega}$, conditions $\mathbf{c}${, unknown future window $\mathbf{X}_{\mathrm{future}}$}}
\KwOut{Sampled future trajectory $\widehat{\mathbf{X}}_{\mathrm{future}}$}
Initialize {$\mathbf{X}_T  \in \mathbb{R}^{(H+F) \times N} \sim \mathcal{N}(0,I)$, $\mathbf{X}_0 = [\mathbf{X}_{\mathrm{past}}, \mathbf{X}_{\mathrm{future}}]$\;}
\For{$t=T$ {\rm to} $1$}{
{ Time series {$t$} embedding}\\
{Positional encoding computed by Eq (\ref{eq:positional_encoding})}\;
Predict noise {with Cross Attention by Eq~(\ref{eq:residual_addition});}
$\hat{\boldsymbol{\epsilon}}_\theta=\epsilon_\theta(\mathbf{X}_t,t,\mathbf{c})$\;
Sample $\mathbf{z} \sim \mathcal{N}(0,I)$ if $t>1$, else $\mathbf{z}=0$\;
Compute one step denoised estimate: $
\mathbf{X}_{t-1}'= \frac{1}{\sqrt{\alpha_t}}\!\left(\mathbf{X}_t - \frac{1-\alpha_t}{\sqrt{1-\bar{\alpha}_t}}\,\hat{\boldsymbol{\epsilon}}_\theta\right) + \sigma_t \mathbf{z} $\;
{Sample $\mathbf{\epsilon} \sim \mathcal{N}(0,I)$ for $\mathbf{X}_{\mathrm{past}}$ if $t>1$, else $\mathbf{\epsilon}=0$\;}
{Add noise to $\mathbf{X}_{\mathrm{0}}$: $\mathbf{X}_\text{obs} 
    = \sqrt{\bar{\alpha}_t}\,\mathbf{X}_0
      + \sqrt{1-\bar{\alpha}_t}\,\boldsymbol{\epsilon}$\;}
{Clamp observed values (imputation step): $\mathbf{\hat{X}}_{t-1}=\boldsymbol{\Omega}\odot\mathbf{X}_{\mathrm{obs}} + (1-\boldsymbol{\Omega})\odot\mathbf{X}_{t-1}'$\;}
      }
Extract $\widehat{\mathbf{X}}_{\mathrm{future}}$ from $\mathbf{X}_0$\;
\label{alg:inference}
\end{algorithm}

\begin{table*}[!t]
    \centering
    \caption{Frequency $f$ (MHz) allocation for mobile operators and technology band used in { the Italian dataset}. Each entry shows  the antenna factor $(\alpha_f)$ in dB m\(^{-1}\) from Keysight N6850A calibration \cite{keysight_n6850a}.}
    \label{tab:operator_frequency_allocation}
    \begin{tabular}{m{0.6cm}<{\raggedleft} m{0.6cm}<{\raggedleft} m{15cm}<{\raggedright}}
        \toprule
        \multirow{7}{*}{\rotatebox{90}{\textbf{Operators}}} 
        & $\mathcal{F}_\mathrm{Iliad}$ 
        & 763.0 (32.8), 1835.0 (31.2), 2150.0 (31.9), 2635.0 (33.2), 3630.0 (36.6) \\
        \cmidrule(lr){2-3}
        & $\mathcal{F}_\mathrm{TIM}$ 
        & 773.0 (32.9), 806.0 (33.3), 935.0 (34.7), 2137.0 (31.8), 2137.5 (31.8), 2662.0 (33.3), 2662.5 (33.3), 3760.0 (36.9), 3468.0 (36.0), 3468.5 (36.0), 3469.0 (36.0), 3568.0 (36.3), 3568.5 (36.3) \\
        \cmidrule(lr){2-3}
        & $\mathcal{F}_\mathrm{VF}$ 
        & 783.0 (33.0), 816.0 (33.4), 945.0 (34.8), 1870.0 (31.0), 2162.0 (31.9), 2162.5 (31.9), 2647.0 (33.2), 2647.5 (33.2), 3680.0 (36.7) \\
        \cmidrule(lr){2-3}
        & $\mathcal{F}_\mathrm{W3}$ 
        & 796.0 (33.2), 955.0 (34.9), 1850.0 (30.9), 2120.0 (31.7), 2585.0 (33.0), 2680.0 (33.4), 3610.0 (36.5) \\
        \midrule\midrule
        \multirow{8}{*}{\rotatebox{90}{\textbf{Technologies}}}
        & $\mathcal{F}_\mathrm{2G}$ 
        & 935.0 (34.7), 945.0 (34.8), 955.0 (34.9) \\
        \cmidrule(lr){2-3}
        & $\mathcal{F}_\mathrm{3G}$ 
        & 2120.0 (31.7), 2137.0 (31.8), 2137.5 (31.8), 2150.0 (31.9), 2162.0 (31.9), 2162.5 (31.9) \\
        \cmidrule(lr){2-3}
        & $\mathcal{F}_\mathrm{4G}$ 
        & 763.0 (32.8), 773.0 (32.9), 783.0 (33.0), 796.0 (33.2), 806.0 (33.3), 816.0 (33.4), 1835.0 (31.2), 1850.0 (30.9), 1870.0 (31.0), 2585.0 (33.0), 2635.0 (33.2), 2647.0 (33.2), 2647.5 (33.2), 2662.0 (33.3), 2662.5 (33.3), 2680.0 (33.4) \\
        \cmidrule(lr){2-3}
        & $\mathcal{F}_\mathrm{5G}$ 
        & 3468.0 (36.0), 3468.5 (36.0), 3469.0 (36.0), 3568.0 (36.3), 3568.5 (36.3), 3610.0 (36.5), 3630.0 (36.6), 3680.0 (36.7), 3760.0 (36.9) \\
        \bottomrule
    \end{tabular}
\end{table*}

\subsection{Model Architecture and Cross-Attention Integration}
\label{subsec:conditional_cross_attention}


We employ a U-Net backbone for the noise-prediction network $\boldsymbol{\epsilon}_{\theta}$. As shown in Fig.~\ref{fig:residual-block}, the U-Net is a deep neural network with multiple residual blocks. The encoder and decoder are symmetric structure, consisting of the same number of residual blocks and additional downsampling or upsampling block.  Each residual block contains convolution to capture local 
{details of the long series, with the input $\mathbf{X_t}$,
condition $\mathbf{c}$,}
and time step $t$. The input multivariate $\mathbf{X}_0$ passes through the U-Net and is processed by residual block to an intermediate feature map $\mathbf{h}$ (with dimensions $C_\mathbf{h} \times h_\mathbf{h} \times w_\mathbf{h}$), while $\mathbf{h}=\mathbf{X}_t$ in the first layer. Here, $C_\mathbf{h}$ is the number of channels in the convolutional layer, $h_\mathbf{h}$ and $w_\mathbf{h}$ are the spatial height and width of the feature map, respectively.

\subsubsection{Residual Block with Time-Step Embedding}
\label{subsec:mv_resblocks}

{
The core of the $\epsilon_\theta$ network is a stack of residual blocks that progressively refine the feature maps. As shown in Fig.~\ref{fig:residual-block}, each residual block  applies normalization, nonlinearity, convolution, and time step conditioning, followed by a residual shortcut.
Given an input feature map $\mathbf{h}$ from the previous layer, the block first applies Group Normalization and a Sigmoid Linear Unit (SiLU) activation, followed by a $3 \times 3$ convolution, i.e.,  \cite{ronneberger2015u}:
\begin{equation}
    \tilde{\mathbf{h}} 
    = \mathrm{Conv}_1\!\bigl(\sigma(\mathrm{GN}(\mathbf{h}))\bigr),
\end{equation}
where $\mathrm{GN}(\cdot)$ denotes GroupNorm, $\sigma(\cdot)$ is the SiLU activation, and $\mathrm{Conv}_1$ is a $3 \times 3$ convolution.

A critical component of the block is the injection of the diffusion step embedding. The global timestep embedding {$\mathbf{t}^e$} is passed through a block-specific linear layer and then added to the intermediate feature map:
\begin{equation}
    \mathbf{h}_i 
    = \tilde{\mathbf{h}} + \mathrm{Linear}_t\!\bigl(\sigma({{\mathbf{t}^e}})\bigr),
\end{equation}
where $\mathrm{Linear}_t$ maps ${{\mathbf{t}^e}} \in \mathbb{R}^{d_{\text{embed}}}$ to the channel dimension of the block, and the result is broadcast across spatial dimensions. This makes the block’s behavior explicitly adaptive to the current diffusion step $t$.

To incorporate the diffusion timestep to the residual block and enrich its effective information, a sinusoidal time embedding is employed to expand its dimension. The timestep $t$ is first mapped into a positional embedding vector $\mathbf{t}^e$ as
\begin{equation}
\begin{aligned}
    \mathbf{t}^e = \text{Concat}\Big( 
    &\left[ \cos(t \cdot \omega_0), \dots, \cos\!\left(t \cdot \omega_{\frac{d}{2}-1}\right) \right],\\
    &\left[ \sin(t \cdot \omega_0), \dots, \sin\!\left(t \cdot \omega_{\frac{d}{2}-1}\right) \right]
    \Big)
\end{aligned}
\end{equation}
For a given dimension index $m$ where $0 \le m < C_\mathbf{h}/2$, the frequencies are defined as follows:
\begin{equation}
    \omega_m = \exp(-\ln(10000) \cdot \frac{2m}{C_\mathbf{h}})
\end{equation}

We integrate condition $\mathbf{c}$ into the feature map using cross-attention operation as  shown below (more details of the cross-attention are given in the following subsection and Fig.~\ref{fig:uv-cross-attention}):
\begin{equation}
\hat{\mathbf{h}} = \mathbf{h}_i + \text{CrossAttention}(\mathbf{h}, \mathbf{c}). \label{eq:residual_addition}
\end{equation}
The updated feature map $\hat {\mathbf{h}}$ is then processed by a second sequence of operations, GroupNorm, SiLU, dropout, and another $3 \times 3$ convolution \cite{ronneberger2015u}, i.e., $
    \mathbf{h}' 
    = \mathrm{Conv}_2\!\bigl(\mathrm{Dropout}(\sigma(\mathrm{GN}(\hat{\mathbf{h}} )))\bigr).
$
Finally, a residual shortcut is applied as \cite{he2016deep}
$
    \mathbf{h}_{\text{out}} = \mathbf{h}' + \mathrm{Shortcut}(\mathbf{h}),
$
where $\mathrm{Shortcut}$ is the identity mapping {for feature mining and convolution operation for up sampling and down sampling}. By stacking these modules {in the encoder}, the series is processed shorter and the convolution channels increase, so that abstract temporal and spatial information are extracted. 
}

{

\subsubsection{Cross-Attention Module} 
 To integrate conditioning information, instead of simple concatenation, we use cross-attention modules at multiple layers in both the encoder and decoder, as depicted in Fig.~\ref{fig:uv-cross-attention}. 
This allows the model to dynamically attend to the most relevant parts of the conditioning information at different stages of processing and across different temporal scales.
 The condition tensor $\mathbf{c}$ (with dimensions $N_f \times d_f$) and the intermediate feature map $\mathbf{h}$
are two inputs of the cross-attention module. {Note that $N_f$ is the sequence length of the condition and $d_f$ is its feature dimension.}

To compute the attention, the 3D feature map $\mathbf{h}$ is first {flattened} into a 2D tensor, $\mathbf{h}_{\text{flat}}$, with shape $(C_h, N_h)$, where $N_h = h_h \cdot w_h$. 
Since cross-attention calculation basically contains linear connection without recurrence, it's important to fully make use of the order of the sequence. Positional encoding adds positional information to $\mathbf{c}$ and $\mathbf{h}_{\text{flat}}$, which helps the module extract temporal and spatial correlations. Consequently, we add positional encoding to $\mathbf{c}$ and $\mathbf{h}_{\text{flat}}$ according to \cite{ref:attention}:
\begin{equation}
\label{eq:positional_encoding}
      \mathbf{c} \revised{\gets} \mathbf{c} + PE(\mathbf{c}) \in \mathbb{R}^{N_f \times d_f}
\end{equation}
\begin{equation}
    \mathbf{h}_{\text{flat}} \leftarrow \mathbf{h}_{\text{flat}}+PE(\mathbf{h}_{\text{flat}}) \in \mathbb{R}^{C_h \times N_h} 
\end{equation}
\begin{equation}
\label{eq:positional_encoding2}
PE(\cdot) =
\begin{cases}
PE_{(p, 2i)} = sin(p/10000^{2i/d_\mathrm{series}}), \\
PE_{(p, 2i+1)} = cos(p/10000^{2i/d_\mathrm{series}}),
\end{cases}
\end{equation}
where $p$ and $i$ denote the position and the dimension index of the input series, respectively, and $d_{\text{series}}$ is the embedding dimension of the sequence.

{The cross-attention module operates between the feature map and the conditional embeddings, allowing the model to explicitly learn their correlations and dynamically modulate its representations based on the provided conditions. Its calculation is based on $QKV$ where the {Queries ($Q$)} are projected from this flattened feature map, while the {Keys ($K$)} and {Values ($V$)} are projected from the conditioning tensor $\mathbf{c}$, i.e.,}

\begin{equation}
\label{eq:QKV}
Q = \mathbf{h}_{\text{flat}} W_{Q}, \quad K = \mathbf{c} W_{K}, \quad V = \mathbf{c} W_{V}
\end{equation}
where $W_Q$, $W_K$, and $W_V$ are learnable weight matrices \cite{ref:latent,huang2025probabilistic}.
The attention matrix $A$ and the final output are calculated using the standard scaled dot-product attention mechanism:
\begin{equation}
\label{eq:cross_attention_calc}
\text{CrossAttention}(\mathbf{h}, \mathbf{c}) =
\text{Softmax}\!\left(\frac{QK^{T}}{\sqrt{e_{k}}}\right)V = AV,
\end{equation}
where $e_{k}$ is the dimension of the keys (and queries), used for scaling. According to \ding{172} in Fig.~\ref{fig:uv-cross-attention}, a single {row} of the attention matrix $A$ represents the set of attention weights that {one specific element} from the flattened feature map $\mathbf{h}_{\text{flat}}$ (represented by a row of $Q$) pays to {all $N_f$ elements} of the condition vector $\mathbf{c}$ (represented by the rows of $K$).
\footnote{
The positional encodings in Eqs.~(\ref{eq:positional_encoding})--(\ref{eq:positional_encoding2}) preserve temporal and sequential order, while the QKV projections in Eq.~(18) learn data-dependent interactions between the EMF feature map and the condition embeddings. 
When certain states are sparse or highly imbalanced, such as holidays or rare seasonal patterns, their corresponding key and value representations may receive fewer gradient updates. 
In such cases, the attention weights in Eq.~(\ref{eq:QKV}) may be more strongly influenced by majority recurring states, such as regular working-hour profiles. Accordingly, the cross-attention module should be interpreted as a flexible conditioning mechanism rather than an explicit solution to contextual imbalance.
}
{\ding{173} in Fig.~\ref{fig:uv-cross-attention} shows a column of the value $V$ that corresponds to a feature map weighted by the attention in $A$ \cite{petit2021u}.}
The output of this operation, $AV$, represents the contextually relevant information from $\mathbf{c}$, weighted according to its relevance to $\mathbf{h}$. This output is then reshaped and projected to match the original feature map's dimensions ($C_h \times h_h \times w_h$). This result is finally added back to the original feature map $\mathbf{h}$ via a residual connection{, which is shown in the previous section}:
\begin{equation*}
\mathbf{h}' = \mathbf{h} + \text{CrossAttention}(\mathbf{h}, \mathbf{c}). 
\end{equation*}
In this context, $\text{CrossAttention}(\cdot)$ in Eq. (\ref{eq:residual_addition}) represents the full module's operation, including the final reshape and projection steps shown in the diagram. According to \ding{173} in Fig.~\ref{fig:uv-cross-attention}, a single {column} of the final output ($AV$) is the new vector representation for that specific feature map element. It is calculated as the {weighted sum of all $N_f$ rows} of the Value matrix $V$. The weights used for this sum are the attention scores from the corresponding row of $A$.

\subsection{EMFusion Training and Inference}
The cross-attention integrated U-Net learns to predict noise in the training phase as Algorithm~\ref{alg:training}. Given the full window $\mathbf{X}_{0}=[\mathbf{X}_{\text{past}}, \mathbf{X}_{\text{future}}]$ and the conditions $\mathbf{c}$, {the first step is to randomly choose a $t \sim \mathcal{U}\{1,\ldots,T\}$ and noise $\boldsymbol{\epsilon} \sim \mathcal{N}(0,I)$ to add noise to the original sequence $\mathbf{X}_{0}$ by using Eq.~\eqref{eq:x_t}: $\mathbf{X}_t=\sqrt{\bar{\alpha}_t}\mathbf{X}_{0}+\sqrt{1-\bar{\alpha}_t}\boldsymbol{\epsilon}$.} The task of the U-Net is to learn the ability to predict $\mathbf{\epsilon}$ given $\mathbf{X}_t, t,\mathbf{c}$. Pass $\mathbf{X}_t$ through U-Net and generate intermediate feature map $\mathbf{h}$. Then embed positional information to $\mathbf{c}$ and $\mathbf{X}_{0}$ and calculate cross-attention between $\mathbf{h}$ and $\mathbf{c}$ when going through cross-attention blocks. Subsequently the output noise $\hat{\boldsymbol{\epsilon}}_\theta$ is the same dimension as $\mathbf{\epsilon}$. By minimizing MSE loss $\mathcal{L}(\theta)=\|\boldsymbol{\epsilon}-\hat{\boldsymbol{\epsilon}}_\theta\|_2^2$, the model updates itself and learn to predict noise based on any $t$. Then the well-trained model participates in the denoising process in the inference phase by predicting noise $\hat{\boldsymbol{\epsilon}}_\theta=\epsilon_\theta(\mathbf{X}_t,t,\mathbf{c})$ at each step $t$ and the conditions are input through cross-attention module, as shown in Algorithm~\ref{alg:inference}.

\subsection{Imputation-Based Sampling at the Inference}
\label{subsec:imputation_sampling}
In practical EMF monitoring, historical measurements are often uneven, partially observed, or corrupted due to sensor outages, data-handling issues, and privacy-related filtering, resulting in time series with substantial gaps and irregular sampling. Such incomplete inputs limit the effectiveness of pure conditional forecasting, which assumes fully reliable past observations. An imputation-based diffusion formulation addresses this limitation by treating both missing past values and all future values as unobserved entries of a single masked sequence. During the DDPM reverse process, the model performs principled generative impainting—reconstructing irregular or missing segments of the past while simultaneously generating future values under the same probabilistic dynamics. Since past and future EMF measurements lie on the same underlying temporal manifold, casting forecasting as structured inpainting enables the model to learn  cross-temporal dependencies across the entire sequence.

To handle both forecasting and imputation within a unified framework, we adopt a masked conditional DDPM that models the full sequence 
{$\mathbf{X}_{\text{0}} = [\mathbf{X}_{\text{past}}, \mathbf{X}_{\text{future}}]$} under partial observations. That is, the model receives a concatenated input sequence {$\mathbf{X}_{\text{0}}$} consisting of a known historical window $\mathbf{X}_{\text{past}}$ and an unknown future window $\mathbf{X}_{\text{future}}$ that is initially masked. The objective is to infer the missing future segment in a manner that is probabilistically consistent with $\mathbf{X}_{\text{past}}$. The algorithmic realization is summarized in \textbf{Algorithm~\ref{alg:inference}}. 

Let $\boldsymbol{\Omega}$ denotes a binary observation mask {(for the i-th element, $\boldsymbol{\Omega}_i=1$ for observed entries and $\boldsymbol{\Omega}_i=0$ for missing ones)}, and let $\mathbf{X}_\text{obs} 
    = \sqrt{\bar{\alpha}_t}\,\mathbf{X}_0
      + \sqrt{1-\bar{\alpha}_t}\,\boldsymbol{\epsilon},
    \quad
    \boldsymbol{\epsilon} \sim \mathcal{N}(0,\mathbf{I}),$ denotes the available measurements. 
As in standard diffusion models, the forward process gradually perturbs the clean sequence $\mathbf{X}_0$ with Gaussian noise as in \eqref{eq:x_t}, 
{while the model learns a masked conditional reverse process, i.e., \eqref{2} still applies by adding conditions on $\mathbf{c}$. The noisy sequence is processed through the conditional U-Net $\boldsymbol{\epsilon}_{\theta}(\mathbf{X}_t, t, \mathbf{c})$ to obtain a one-step denoised estimate of the entire trajectory, denoted as $\hat{\mathbf{X}}_{t-1}$.}
After each denoising step, the observed entries are clamped to their known values to ensure consistency:
\begin{equation}
\mathbf{\hat{X}}_{t-1} = \mathbf{\boldsymbol{\Omega}} \odot \mathbf{X}_{\text{obs}} + (1 - \mathbf{\boldsymbol{\Omega}})\odot \mathbf{{X}}'_{t-1},
\end{equation}
where $\hat{X}_{t-1}$ denotes the model’s predicted denoised sample at timestep $t-1$.  
\begin{equation}
    \hat{\mathbf{X}}_{t-1}= \frac{1}{\sqrt{\alpha_t}}\!\left(\mathbf{X}_t - \frac{1-\alpha_t}{\sqrt{1-\bar{\alpha}_t}}\,\hat{\boldsymbol{\epsilon}}_\theta\right) + \sigma_t \mathbf{z} 
\end{equation}

{
After $T$ iterations of the denoising process, a predicted sequence $\widehat{\mathbf{X}}_\mathrm{future}$ can be obtained from $\mathbf{\hat{X}}_0$. 
}
This mechanism enables the DM to impute with past values and generate  future trajectories within the same sampling procedure.  
The training objective follows the standard noise-prediction loss as in \eqref{eq:loss_training} with masked conditioning. If we sample $\hat{N}$ times, a set of $\hat{N}$ predicted sequences $\mathbf{S}=\{\hat{\mathbf{X}}_\mathrm{future}^i\}_{i=1,\cdots, \hat{N}}$ is ready for probabilistic interval construction which is explained next.}

\subsection{Probabilistic Interval Construction}
\label{subsec:mv_intervals}
{Once EMFusion generates an ensemble of $\hat{N}$ future EMF trajectories 
$\mathbf{S}=\{\hat{\mathbf{X}}_{\mathrm{future}}^i\}_{i=1}^{\hat{N}} 
\in \mathbb{R}^{\hat{N} \times F}$ for each monitored carrier (e.g., individual 2G, 3G, 4G, and 5G carriers across multiple operators), 
the samples at forecast horizon $k \in \{1,\dots,F\}$ are given by
$
\mathbf{Z}_k = \{ z_k^{1}, \dots, z_k^{\hat{N}} \}.
$
These samples are interpreted as Monte-Carlo draws from the learned 
conditional predictive distribution. A smooth estimate of the predictive 
density at horizon $k$ is obtained via Kernel Density Estimation (KDE) as shown below:
\begin{equation}
\hat f_k(z)
=
\frac{1}{\hat N s_p}
\sum_{i=1}^{\hat N}
K\!\left(\frac{z - z_k^{i}}{s_p}\right),
\end{equation}
where $K(\cdot)$ denotes the Gaussian kernel function and $s_p$ is the 
bandwidth parameter controlling the bias--variance trade-off.
The corresponding cumulative distribution function is
$
\hat F_k(z)
=
\int_{-\infty}^{z}
\hat f_k(t)\, dt.
$
For a target confidence level $\gamma$, define
$
\underline{\alpha} = \frac{1-\gamma}{2},
\qquad
\overline{\alpha} = 1 - \underline{\alpha}.
$
The prediction interval based on the estimated density is given by:
\begin{equation}
PI_\gamma
=
\left[
\hat F_k^{-1}(\underline{\alpha}),
\hat F_k^{-1}(\overline{\alpha})
\right].
\end{equation}
In practical implementation, instead of explicitly evaluating the KDE and 
numerically inverting the CDF, the required quantiles are computed directly 
from the ordered ensemble samples. Let
$
z_k^{(1)} \le \dots \le z_k^{(\hat N)}
$
denotes the order statistics of $\mathbf{Z}_k$. Let us define
$
\ell = \left\lceil \underline{\alpha}\hat N \right\rceil,
\quad
u = \left\lceil \overline{\alpha}\hat N \right\rceil.
$
The empirical prediction interval is then given by
$
PI_\gamma
=
[L_k, U_k]
=
\left[
z_k^{(\ell)},
z_k^{(u)}
\right].
$
As the ensemble size $\hat N$ increases, the empirical quantiles 
converge to those of the underlying  distribution, 
ensuring statistically consistent and efficient uncertainty estimation.}

\subsection{Univariate Forecasting: A Special Case}
\begin{figure}[t]
    \centering
    \includegraphics[width=0.8\linewidth]{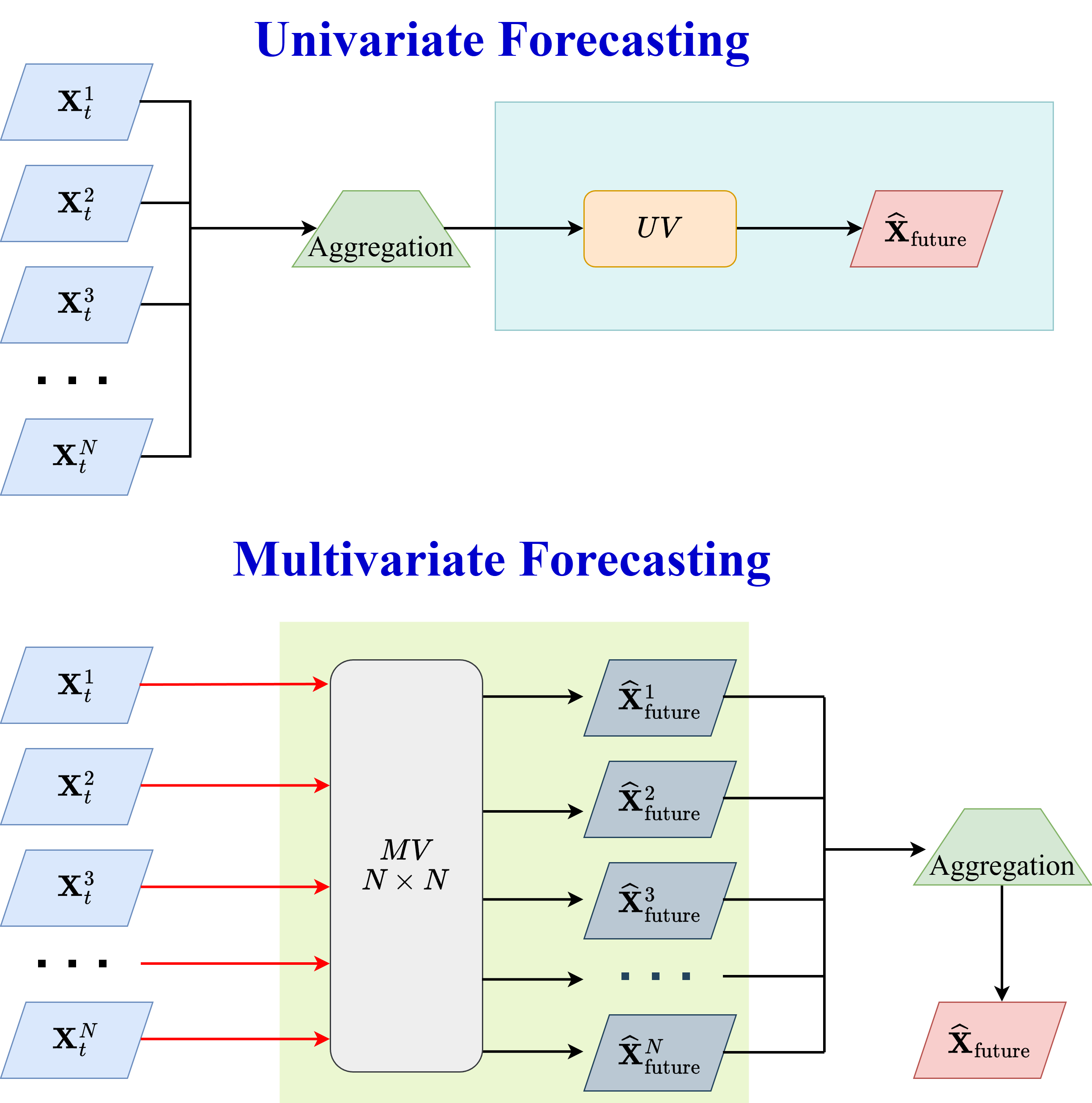}
    \caption{Comparison of univariate (UV) and multivariate (MV) forecasting strategies for  EMF time-series across $N$ features.
}
    \label{fig:uv_mv}
\end{figure}
{As illustrated in Fig.~\ref{fig:uv_mv}, 
Univariate forecasting illustrates the {aggregate-then-learn} paradigm, where multiple time series are first aggregated before applying the forecasting model.  
Multivariate forecasting the {learn-then-aggregate} paradigm, where forecasts are generated for each frequency series individually and then aggregated. 
}
In the EMFusion framework, the main differences between the univariate and multivariate instantiations of EMFusion are confined to the input dimensionality and the associated architectural hyperparameters. In the multivariate setting ($N > 1$), the input $\mathbf{X}_0$ is treated as a 2-D tensor, processed via 2-D convolutional layers to simultaneously capture temporal evolution and inter-frequency correlations. 
Conversely, the univariate configuration ($N=1$) utilizes 1-D convolutions, focusing exclusively on temporal dynamics. In the univariate setting, the model receives a single narrow-band EMF time series and the input tensor is of shape $(B, H, 1)$, where $B$ is the batch size and $H$ is the look-back length, thus we employ 1-D convolutions. In the multivariate setting, the input at each time step lead to an input tensor of shape $(B, H, N)$ and 2-D convolutions  acting jointly on time and frequency. When computing cross-attention, the univariate  uses a single hidden channel $h_h = 1$ in the value projection (effectively attending only along time), whereas the multivariate sets $h_h > 1$ so that attention can propagate information both temporally and across bands.

\section{Data Specifications and Analysis}
\label{sec:expierment_data_visualization}
This section details dataset specifications including the measurement set-up, received power (dBm) to electric-field (Volts per meter) conversion, and the external conditioning factors.
\begin{figure}
    \centering
    \includegraphics[width=\columnwidth]{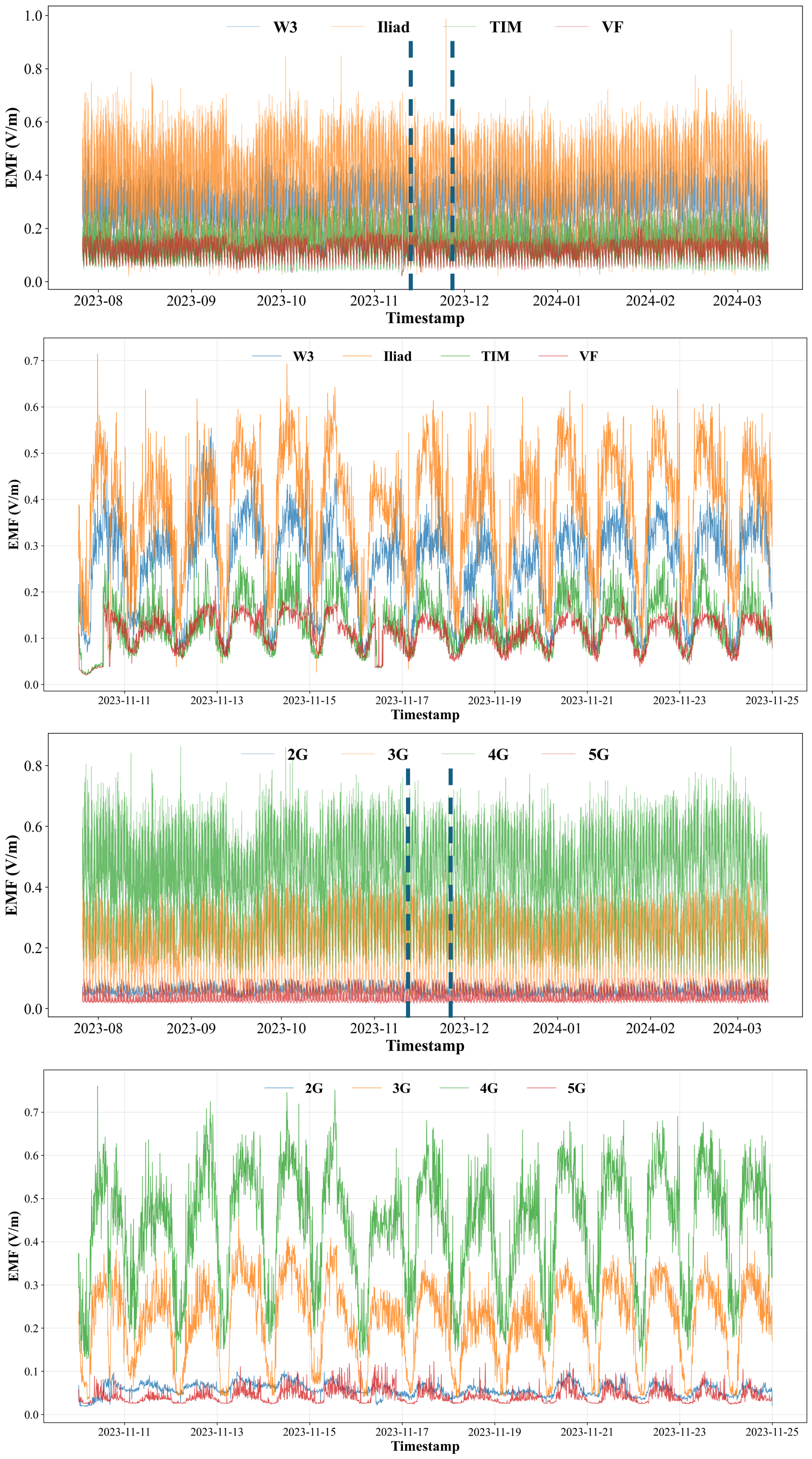}
    \caption{EMF levels in {Italian dataset}, as a function of network operators (1,2) and cellular technologies (3,4). The plots (1,3) display the 8-month trend (Aug 2023 – Mar 2024), while the plots (2,4) show enlarged 14-day daily trend.}
    \label{fig:emf_operator}
\end{figure}
The top two plots in Fig.~\ref{fig:emf_operator} show the time evolution\footnote{The data were collected at the University Hospital Tor Vergata in Rome, Italy, from 2023-07-26 16:49:02 to 2024-03-11 05:08:28. } of operator–specific EMF levels in the Italian dataset.  The operator-wise difference in EMF levels is notable. Iliad systematically dominates the total exposure, with a typical peak at $0.8$ v/m. W3 provides the second-largest contribution on EMF, while TIM and VF remain consistently lower, with most values below $0.3$~V/m. 
{This suggests antenna being placed in the University hospital or a configuration that produced a stronger signal at the specific point of measurement in favour of Iliad. Moreover, while the other operators display broadly similar deployment and load
patterns, Iliad follows a more independent deployment and load profile, with a
narrower spectrum allocation in the mid-band 5G frequencies compared to the other
operators.  }

\begin{figure}[t]
    \centering
    \includegraphics[width=\linewidth]{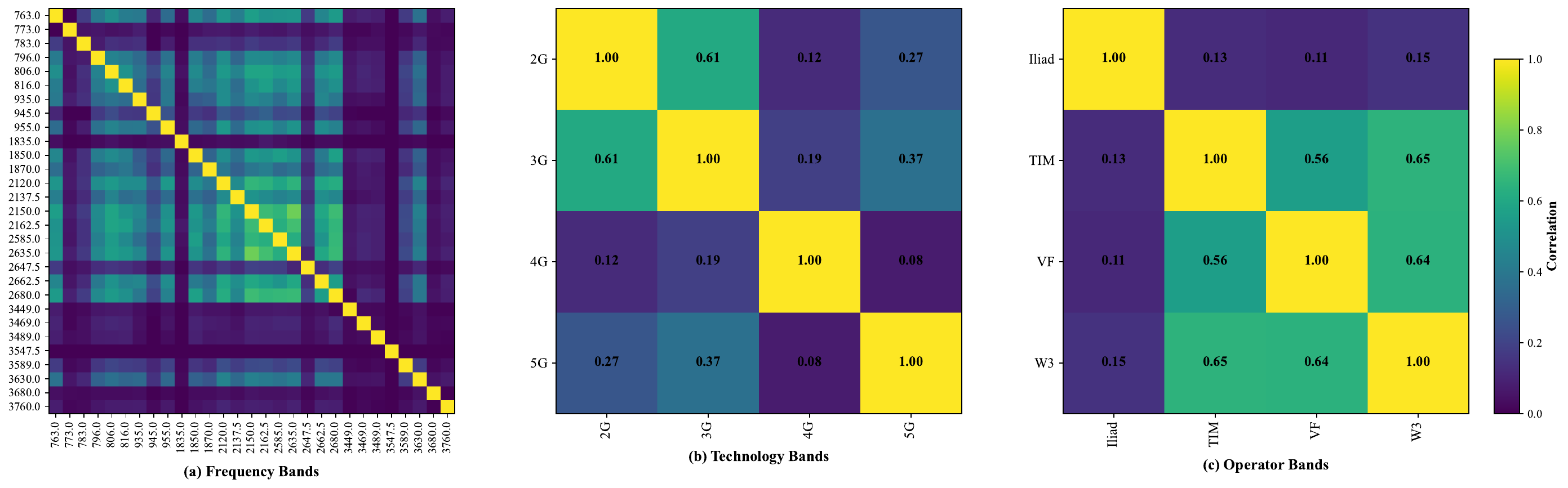}
    \caption{Correlation maps of EMF levels for (a) frequency bands, (b) cellular technologies, and (c) network operators.}
    \label{fig:three_heatmaps}
\end{figure}

On the other hand, the bottom two plots in Fig.~\ref{fig:emf_operator} depict that the EMF level from 4G  clearly dominates the total EMF levels from 2G, 3G, or 5G technologies,
both in terms of average level and short–term variability, with typical peaks
reaching $0.8$~V/m. The 3G contribution is noticeably smaller but still
non-negligible, while 2G and 5G produce the lowest fields, generally below
$0.1$~V/m except for a few isolated peaks. This trend is likely due to the sparse deployment of both 2G and 5G technologies and widespread deployment of 4G at the time of measurements. 
Both operator-wise and technology-wise display the recurrent daily cycle and stable behaviour over the months.
{According to Fig.~\ref{fig:three_heatmaps}, 4G is only weakly correlated with the
other technologies ($\rho \leq 0.19$), whereas 5G exhibits a moderate correlation
with 2G and 3G. This behavior is consistent
with the technological innovations introduced in 4G and 5G with respect to legacy
generations, such as active antennas, beamforming, MIMO, and time-division
duplexing. 
}

\subsection{Dataset Specifications}
The received power measurements were collected continuously from July 26, 2023 to March 10, 2024 across 29 frequency channels spanning from 9 kHz - 6GHz. The dataset is then organized according to the predefined frequency bands of major Italian network operators and the corresponding cellular technologies. Table~\ref{tab:operator_frequency_allocation} summarizes the frequency allocations used by each operator and technology.

\subsubsection{EMF Monitoring Set-up} The EMF monitoring chain couples an omnidirectional antenna, an Anritsu MS27102A remote spectrum-monitoring unit, and the ARPA-Lazio \textit{Search24} control software \cite{chiaraviglio2022six}.  
The Keysight N6850A passive antenna \cite{keysight_n6850a} is connected to the portable spectrum analyzer via a 1.5 m low-loss coaxial cable. The antenna delivers uniform gain across all licensed cellular bands up to the 3.6 GHz.  The weather-proof MS27102A sweeps 9 kHz–6 GHz continuously and, thanks to its internal pre-selector, outputs real-time narrow-band channel power while operating unattended on a rooftop site~\cite{ms27102a}.  
\textit{Search24} \cite{search24} scripts enable streaming the data to a local computer over a dedicated Ethernet link.  

\subsubsection{Conversion of Received Power to EMF}
\label{sec:power2emf}
The  received power for a given frequency channel $f$ at the output of the
 antenna is measured in $\text{dBm}$.  For a given frequency $f$, the received power is then converted into  the  electric-field strength (measured in V/m)  as given below \cite{chiaraviglio2021health}:
\begin{equation}
  E_f  = \sqrt{\frac{P_{\mathrm{r}}(f)\,Z_{0}} {A_{\mathrm{e}}(f)}} 
  =\sqrt{\frac{10^{\frac{P_{\mathrm{dBm}}(f){-}30}{10}}Z_{0}}{ A_{\mathrm{e}}(f)}}
  \label{eq:e_field}
\end{equation}
where  $Z_{0}\!\approx\! 376.73~\Omega$, which is the intrinsic (wave) impedance of free space.
For a measurement antenna with linear gain $G_f$, the  effective aperture  $A_{\mathrm{e}}(f)$ is defined as follows:
\begin{equation}
 A_{\mathrm{e}}(f)
  = \frac{G_f\lambda_f^{2}}{4\pi},
  \quad
  \lambda_f = \frac{c}{f},
  \quad
  G_f = \left( \frac{A_{GC}}{\lambda_f \cdot 10^{\alpha_f/20}} \right)^2
  \label{eq:aperture}
\end{equation}
where $c$ is the speed of light. $A_{GC}$ is antenna-gain calibration constant. The antenna factor $\alpha_f$ changes as a function of frequency $f$ as in  \cite{chiaraviglio2022six,keysight_n6850a_antfac}.

\subsubsection{External Conditions}  
User activity patterns, driven by work schedules and lifestyle habits, directly influence device usage and consequently the received power and EMF levels. To account for these temporal variations, calendar-based contextual information is incorporated as an external conditioning signal for EMF prediction.
For each observation with time stamp $t$, three features were appended:
\begin{itemize} 
  \item \textbf{Italy\_WorkingDay} $\in \{0,1\}$: indicates whether $t$ falls on a weekday (Monday–Friday) {and} is {not} listed in the Italian public–holiday calendar.\footnote{The holiday list covers national and widely observed regional holidays for 2023–2024, including Easter, Liberation Day, Labour Day, Republic Day, Ferragosto, All Saints’ Day, Immaculate Conception, and the Christmas–New Year period. It also includes the university closure date and no teaching activities dates for Università degli Studi di Roma Tor Vergata.\cite{TorVergataAcademicCalendarPage2025} }
  \item \textbf{Italy\_WorkingHour} $\in \{0,1\}$: equals 1 when the local clock time lies inside the standard business window $[09{:}00,17{:}00)$ and 0 otherwise.
  \item \textbf{Italy\_Season} $\in \{1,2,3,4\}$: a flag assigned based on the astronomical equinox–solstice boundaries, i.e.,  1) \textit{Spring}: March 21 to June 20; 2) \textit{Summer}: June 21 to September 22; 3) \textit{Autumn}: September 23 to December 20; and 4) \textit{Winter}: December 21 to March 20 in the next year.
 {
\item \textbf{Italy\_Multi} $\in \{1,2,3,4\} \times \{0,1\}$: a two-dimensional multi-condition feature obtained by concatenating {Italy\_Season} and {Italy\_WorkingDay}.
 }

\end{itemize}

\subsection{EMF Classification and Correlation Analysis}
From Table~\ref{tab:operator_frequency_allocation}, the aggregate EMF associated with any given operator or cellular technology can be obtained as follows: 
\begin{equation}
    E_{\mathcal{F}_\chi}(t)
    = \biggl( \sum_{f \in \mathcal{F}_\chi} E^2_f \biggr)^{1/2},
\end{equation}
where $\mathcal{F}_\chi \in \{\mathrm{2G}, \mathrm{3G}, \mathrm{4G}, \mathrm{5G}, \mathrm{TIM}, \mathrm{VF}, \mathrm{W3}, \mathrm{Iliad}\}.
$
As an illustration, the 2G band corresponds to the frequency set
$
\mathcal{F}_\mathrm{2G} = \{935.0, 945.0, 955.0\}\,\text{MHz},
$
so the aggregate 2G EMF can be computed as
$
    E_\mathrm{2G}(t)
    = \sqrt{E^2_{935.0} + E^2_{945.0} + E^2_{955.0}}.
$

 Fig. ~\ref{fig:three_heatmaps}(a) depicts the correlation of EMF among all frequencies. We note that the EMF levels across the bands (e.g., 700–900 MHz, 1800–2100 MHz, 2.6 GHz, and 3.4–3.8 GHz) tend to be more strongly correlated with each other indicating that exposure dynamics are largely driven by co-located carriers within the same spectral neighbourhood.

 Fig.~\ref{fig:three_heatmaps}(b) depicts the correlation of EMF observed among the cellular technologies in the {Italian dataset}. 2G and 3G exhibit the strongest off-diagonal correlation ($\rho \approx 0.61$), reflecting similar temporal usage patterns of legacy technologies at this location. In contrast, 4G is only weakly correlated with the other technologies ($\rho \le 0.19$), while 5G shows moderate correlation with 2G and 3G ($\rho \approx 0.27$–$0.37$).

The EMF correlation map (Fig.~\ref{fig:three_heatmaps}(c)) reveals a high correlation among TIM, VF, and W3, with pairwise correlations in the range $\rho \approx 0.56$–$0.65$, while Iliad remains only weakly correlated with the other operators ($\rho \approx 0.11$–$0.15$). This indicates that the EMF exposure patterns of the three incumbent operators are highly synchronized at the hospital site, whereas Iliad follows a more independent deployment and load profile.

\begin{table*}[t]
\centering
\small
\caption{A comparative analysis of univariate forecasting performance of EMFusion with imputation without external conditions. Metrics are ordered as CRPS, ND, NRMSE, MAPE, and PICP. Lower is better for CRPS, ND, NRMSE, MAPE; higher is better for PICP. \textbf{Bold} = best in column; \underline{underline} = second best in column.}
\label{tab:uv_mv_common_metrics_5m}

{
\renewcommand{\arraystretch}{1.3}
\resizebox{\linewidth}{!}{%
\setlength{\tabcolsep}{0.2em}
\begin{tabular}{l|*{20}{c}}
\toprule
\multicolumn{21}{c}{\textbf{Operator-wise Performance}} \\
\midrule
 & \multicolumn{5}{c}{\textbf{Iliad}} & \multicolumn{5}{c}{\textbf{TIM}} & \multicolumn{5}{c}{\textbf{VF}} & \multicolumn{5}{c}{\textbf{W3}} \\
\cmidrule(lr){2-6} \cmidrule(lr){7-11} \cmidrule(lr){12-16} \cmidrule(lr){17-21}
Model & CRPS & ND & NRMSE & MAPE & PICP
      & CRPS & ND & NRMSE & MAPE & PICP
      & CRPS & ND & NRMSE & MAPE & PICP
      & CRPS & ND & NRMSE & MAPE & PICP \\
\midrule
NF &0.0224 & 0.1901 & 0.2271 & 22.0056 & 69.36 & 0.0110 & 0.2674 & 0.3344 & 30.5366 & \underline{65.99} & 0.0057 & 0.1505 & 0.1899 & 16.3322 & \underline{69.26} & \underline{0.0134} & 0.1732 & 0.2126 & 19.5926 & \textbf{72.44} \\
IQLSTM &\underline{0.0208} & 0.1742 & 0.2088 & 22.0082 & \underline{69.61} & 0.0098 & 0.2272 & 0.2905 & 23.7362 & 65.43 & 0.0060 & 0.1484 & 0.2158 & 16.5989 & 64.66 & 0.0146 & 0.1742 & 0.2179 & 19.3497 & \underline{67.87} \\
DDPM &0.0281 & 0.2266 & 0.2752 & 24.7179 & \textbf{75.38} & 0.0131 & 0.3078 & 0.4063 & 35.0737 & \textbf{76.78} & 0.0067 & 0.1741 & 0.2173 & 19.2858 & \textbf{90.22} & 0.0338 & 0.3945 & 0.4217 & 42.9547 & 28.02 \\
VAE &0.0255 & 0.2040 & 0.2424 & 24.0345 & 54.86 & 0.0121 & 0.2880 & 0.3504 & 33.7195 & 53.63 & 0.0063 & 0.1630 & 0.2040 & 17.8781 & 58.19 & 0.0146 & 0.1842 & 0.2236 & 21.5268 & 66.11 \\
LSTM-Dropout &0.0311 & 0.2298 & 0.2626 & 32.2884 & 33.07 & 0.0137 & 0.2944 & 0.3334 & 38.0923 & 31.38 & 0.0072 & 0.1687 & 0.2004 & 20.5388 & 31.19 & 0.0232 & 0.2515 & 0.2747 & 34.6254 & 31.17 \\
WGAN &0.0237 & 0.1856 & 0.2216 & 21.6153 & 50.18 & 0.0207 & 0.3413 & 0.3809 & 44.7403 & 0.00 & \underline{0.0055} & \underline{0.1419} & \textbf{0.1780} & 15.6357 & 60.28 & 0.0138 & 0.1741 & 0.2096 & 20.5489 & 62.22 \\
TimeGrad (29) &0.0329 & 0.3405 & 0.8942 & 89.2536 & --- & \underline{0.0094} & 0.4230 & 1.3310 & 52.4586 & --- & 0.0084 & 0.2997 & 1.0983 & 29.7099 & --- & 0.0203 & 0.3184 & 1.0497 & 48.2214 & --- \\
TimeGrad (8)v2 &0.0552 & 0.1790 & 0.5465 & 26.7691 & --- & 0.0271 & 0.3545 & 1.1546 & 34.5811 & --- & 0.0201 & 0.1785 & 1.0200 & \underline{15.6345} & --- & 0.0367 & 0.2124 & 0.7484 & 18.9919 & --- \\
EMForecaster &0.0243 & \underline{0.1706} & \underline{0.2019} & \underline{21.5334} & 34.50 & 0.0103 & \underline{0.2042} & \underline{0.2499} & \underline{22.5550} & 29.67 & 0.0061 & \underline{0.1419} & 0.2031 & 16.0613 & 44.94 & 0.0147 & \underline{0.1585} & \underline{0.1937} & \underline{18.5124} & 37.61 \\
\textbf{EMFusion} &\textbf{0.0169} & \textbf{0.1304} & \textbf{0.1772} & \textbf{14.2795} & 50.40 & \textbf{0.0067} & \textbf{0.1433} & \textbf{0.2143} & \textbf{14.1847} & 58.51 & \textbf{0.0043} & \textbf{0.0997} & \underline{0.1781} & \textbf{11.0273} & 52.25 & \textbf{0.0099} & \textbf{0.1132} & \textbf{0.1639} & \textbf{12.1131} & 52.19 \\
\bottomrule
\end{tabular}
}

\vspace{0.6em}

\resizebox{\linewidth}{!}{%
\setlength{\tabcolsep}{0.2em}
\begin{tabular}{l|*{20}{c}}
\toprule
\multicolumn{21}{c}{\textbf{Technology-band-wise Performance}} \\
\midrule
& \multicolumn{5}{c}{\textbf{2G}} & \multicolumn{5}{c}{\textbf{3G}} & \multicolumn{5}{c}{\textbf{4G}} & \multicolumn{5}{c}{\textbf{5G}} \\
\cmidrule(lr){2-6} \cmidrule(lr){7-11} \cmidrule(lr){12-16} \cmidrule(lr){17-21}
Model & CRPS & ND & NRMSE & MAPE & PICP
      & CRPS & ND & NRMSE & MAPE & PICP
      & CRPS & ND & NRMSE & MAPE & PICP
      & CRPS & ND & NRMSE & MAPE & PICP \\
\midrule
NF &0.0028 & 0.1416 & 0.1773 & 14.3473 & 59.79 & 0.0150 & 0.2153 & 0.2534 & 24.2409 & 62.59 & 0.0217 & 0.1637 & 0.1983 & 18.1657 & \underline{69.23} & \underline{0.0024} & 0.2103 & 0.3111 & 18.5137 & \textbf{72.20} \\
IQLSTM &\underline{0.0027} & \underline{0.1342} & 0.1708 & \underline{13.7036} & 56.70 & \underline{0.0134} & 0.1841 & 0.2256 & 20.9445 & \underline{63.17} & 0.0210 & \underline{0.1488} & \underline{0.1827} & \underline{17.1465} & 66.22 & 0.0028 & 0.1901 & \underline{0.2994} & \underline{18.1656} & 66.50 \\
DDPM &0.0029 & 0.1462 & 0.1858 & 15.0905 & \textbf{82.02} & 0.0150 & 0.2132 & 0.2536 & 26.5512 & \textbf{76.83} & 0.0267 & 0.1890 & 0.2345 & 19.6537 & \textbf{74.21} & 0.0031 & 0.2536 & 0.3570 & 24.2072 & \underline{71.12} \\
VAE &0.0028 & 0.1429 & 0.1786 & 14.3644 & \underline{60.92} & 0.0168 & 0.2227 & 0.2606 & 26.1413 & 45.59 & 0.0237 & 0.1698 & 0.2051 & 19.2985 & 58.84 & 0.0028 & 0.2298 & 0.3246 & 21.2987 & 57.09 \\
LSTM-Dropout &0.0033 & 0.1388 & \underline{0.1699} & 14.4200 & 32.48 & 0.0196 & 0.2728 & 0.2991 & 44.0706 & 48.00 & 0.0294 & 0.1994 & 0.2273 & 24.2642 & 39.79 & 0.0038 & 0.2742 & 0.3224 & 28.7500 & 14.29 \\
WGAN &0.0028 & 0.1374 & 0.1747 & 13.7520 & 53.91 & 0.0146 & 0.2003 & 0.2339 & 23.7100 & 54.25 & 0.0224 & 0.1556 & 0.1903 & 17.8671 & 51.53 & 0.0046 & 0.2674 & 0.3303 & 26.9596 & 0.00 \\
TimeGrad (29) &0.0059 & 0.3761 & 2.2558 & 38.7363 & --- & 0.0267 & 0.3133 & 0.9559 & 61.6179 & --- & 0.0190 & 0.3375 & 0.9451 & 56.1075 & --- & 0.0051 & 0.3912 & 0.9879 & 40.6475 & --- \\
TimeGrad (8)v2 &0.0072 & 0.1746 & 1.9840 & 16.9011 & --- & 0.0339 & 0.2040 & 0.6425 & 24.7493 & --- & 0.0584 & 0.1888 & 0.6344 & 21.9084 & --- & 0.0111 & 0.3741 & 1.0243 & 21.8487 & --- \\
EMForecaster &0.0104 & 0.1682 & 0.2156 & 19.0429 & 37.40 & 0.0139 & \underline{0.1688} & \underline{0.2122} & \underline{19.6655} & 36.68 & \textbf{0.0134} & 0.1661 & 0.2102 & 19.3257 & 37.34 & 0.0128 & \underline{0.1802} & \textbf{0.2262} & 20.6762 & 34.69 \\
\textbf{EMFusion} &\textbf{0.0016} & \textbf{0.0821} & \textbf{0.1135} & \textbf{8.3460} & 59.64 & \textbf{0.0085} & \textbf{0.1153} & \textbf{0.1564} & \textbf{13.1791} & 52.88 & \underline{0.0169} & \textbf{0.1109} & \textbf{0.1495} & \textbf{11.8418} & 55.69 & \textbf{0.0019} & \textbf{0.1518} & 0.4311 & \textbf{11.5743} & 66.95 \\
\bottomrule
\end{tabular}
}
}
\end{table*}

\begin{figure}[t]
    \centering
    \includegraphics[width=\linewidth]{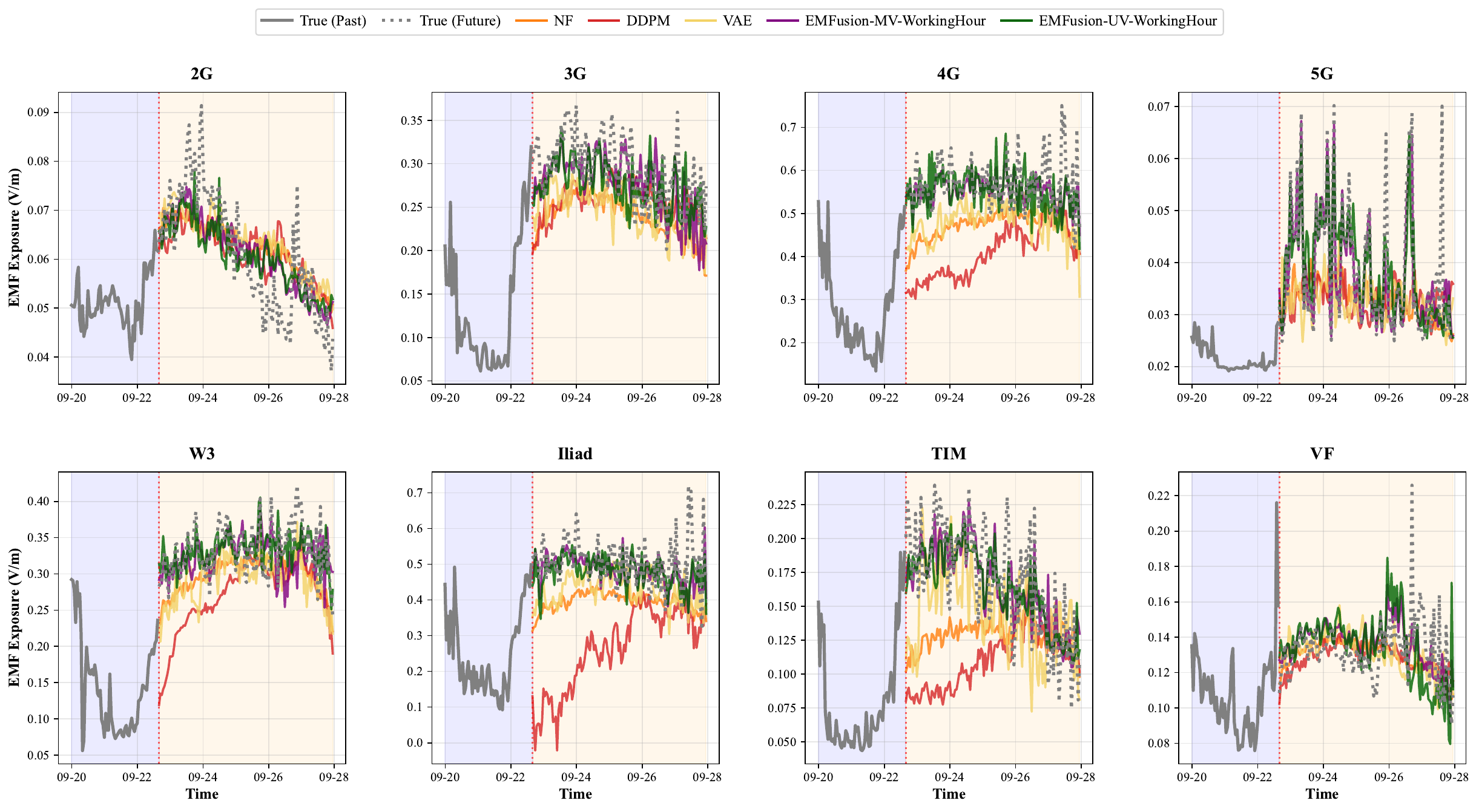}
    \caption{Comparison of EMF Forecasts in Validation Set Across Technology Bands (Top) and Operators (Bottom)
}
    \label{fig:8in1_baseliness}
\end{figure}

\begin{table*}[t]
\centering
\small
\caption{Forecasting performance of EMFusion  without external conditions. Metrics are ordered as CRPS, ND, NRMSE, MAPE, and PICP. Lower is better for CRPS, ND, NRMSE, MAPE; higher is better for PICP. \textbf{Bold} = best in row; \underline{underline} = second best in row. PICP values lower than 10 are considered abnormal and replaced with ``--''.}
\label{tab:forecasting_comparison}
\resizebox{\linewidth}{!}{
\setlength{\tabcolsep}{0.8em}
\begin{tabular}{lll|ccccc|ccccc}
\toprule
\multicolumn{3}{c}{\textbf{}} &
\multicolumn{5}{c|}{\textbf{Univariate}} &
\multicolumn{5}{c}{\textbf{Multivariate}} \\
\cmidrule(lr){4-8} \cmidrule(lr){9-13}
\multicolumn{3}{c}{\textbf{Metric}} &
\textbf{\shortstack{EMFusion \\ (Uncond.)}} &
\textbf{\shortstack{EMFusion \\ (Seasons)}} &
\textbf{\shortstack{EMFusion \\ (Workingday)}} &
\textbf{\shortstack{EMFusion \\ (WorkingHour)}} &
\textbf{\shortstack{EMFusion \\ (multi)}} &
\textbf{\shortstack{EMFusion \\ (Uncond.)}} &
\textbf{\shortstack{EMFusion \\ (Seasons)}} &
\textbf{\shortstack{EMFusion \\ (Workingday)}} &
\textbf{\shortstack{EMFusion \\ (WorkingHour)}} &
\textbf{\shortstack{EMFusion \\ (multi)}} \\
\midrule\midrule

\multirow{20}{*}{\rotatebox{90}{\textbf{Operator}}}

& \multirow{5}{*}{W3}
& CRPS &0.0099 & 0.0111 & 0.0103 & \underline{0.0097} & 0.0193 & 0.1096 & 0.1095 & 0.1091 & \textbf{0.0086} & 0.0365 \\
& & ND &\underline{0.1108} & 0.1265 & 0.1230 & 0.1156 & 0.2179 & 0.8993 & 0.8950 & 0.8935 & \textbf{0.0946} & 0.3300 \\
& & NRMSE &\underline{0.1601} & 0.1708 & 0.1715 & 0.1669 & 0.2673 & 0.9086 & 0.9041 & 0.9028 & \textbf{0.1467} & 0.3960 \\
& & MAPE &\underline{11.79} & 14.13 & 13.77 & 12.60 & 25.30 & 88.29 & 87.87 & 87.69 & \textbf{9.89} & 38.06 \\
& & PICP &50.49 & 48.91 & \underline{54.78} & \textbf{60.26} & 43.41 & -- & -- & -- & 52.69 & 24.88 \\

\cmidrule(lr){2-13}

& \multirow{5}{*}{Iliad}
& CRPS &0.0164 & 0.0173 & 0.0167 & \underline{0.0153} & 0.0273 & 0.0724 & 0.0724 & 0.0680 & \textbf{0.0120} & 0.0435 \\
& & ND &0.1215 & 0.1334 & 0.1319 & \underline{0.1211} & 0.1975 & 0.4382 & 0.4382 & 0.4041 & \textbf{0.0903} & 0.2759 \\
& & NRMSE &0.1744 & 0.1744 & 0.1795 & \underline{0.1724} & 0.2380 & 0.4819 & 0.4819 & 0.4419 & \textbf{0.1469} & 0.3348 \\
& & MAPE &14.05 & 15.94 & 15.05 & \underline{13.93} & 26.39 & 40.38 & 40.38 & 38.15 & \textbf{10.40} & 31.89 \\
& & PICP &50.15 & 52.66 & 53.51 & \underline{56.40} & 39.08 & 11.37 & 11.37 & 10.76 & \textbf{63.02} & 27.86 \\

\cmidrule(lr){2-13}

& \multirow{5}{*}{TIM}
& CRPS &0.0066 & 0.0072 & 0.0066 & \textbf{0.0063} & 0.0122 & 0.0475 & 0.0475 & 0.0482 & \underline{0.0065} & 0.0215 \\
& & ND &\underline{0.1381} & 0.1548 & 0.1522 & 0.1466 & 0.2572 & 0.7524 & 0.7524 & 0.7579 & \textbf{0.1374} & 0.3891 \\
& & NRMSE &\textbf{0.2094} & 0.2180 & 0.2254 & 0.2184 & 0.3219 & 0.7756 & 0.7756 & 0.7810 & \underline{0.2126} & 0.4356 \\
& & MAPE &\textbf{13.73} & 16.20 & 15.91 & 15.02 & 25.63 & 72.80 & 72.80 & 73.31 & \underline{13.81} & 43.49 \\
& & PICP &57.05 & 51.81 & \underline{57.11} & \textbf{57.68} & 39.58 & -- & -- & -- & 54.97 & 25.31 \\

\cmidrule(lr){2-13}

& \multirow{5}{*}{VF}
& CRPS &\underline{0.0043} & 0.0046 & 0.0044 & \underline{0.0043} & 0.0077 & 0.0349 & 0.0349 & 0.0331 & \textbf{0.0036} & 0.0122 \\
& & ND &\underline{0.1005} & 0.1118 & 0.1078 & 0.1043 & 0.1806 & 0.6308 & 0.6308 & 0.6061 & \textbf{0.0826} & 0.2498 \\
& & NRMSE &\underline{0.1762} & 0.1878 & 0.1872 & 0.1862 & 0.2445 & 0.6438 & 0.6438 & 0.6210 & \textbf{0.1571} & 0.3039 \\
& & MAPE &\underline{11.46} & 12.35 & 12.15 & 11.74 & 19.34 & 62.48 & 62.48 & 59.82 & \textbf{9.67} & 25.23 \\
& & PICP &51.04 & 54.16 & 55.79 & \textbf{58.90} & 44.61 & -- & -- & -- & \underline{56.56} & 29.63 \\

\midrule

\multirow{20}{*}{\rotatebox{90}{\textbf{Technology Band}}}

& \multirow{5}{*}{2G}
& CRPS &0.0070 & 0.0076 & 0.0071 & \underline{0.0068} & 0.0131 & 0.0640 & 0.0639 & 0.0635 & \textbf{0.0062} & 0.0234 \\
& & ND &\underline{0.1165} & 0.1310 & 0.1276 & 0.1222 & 0.2186 & 0.7609 & 0.7594 & 0.7525 & \textbf{0.1049} & 0.2922 \\
& & NRMSE &\underline{0.1819} & 0.1922 & 0.1947 & 0.1905 & 0.2779 & 0.7760 & 0.7745 & 0.7683 & \textbf{0.1721} & 0.3542 \\
& & MAPE &\underline{12.33} & 14.22 & 13.94 & 13.12 & 23.42 & 74.52 & 74.38 & 73.61 & \textbf{11.13} & 29.57 \\
& & PICP &52.86 & 51.62 & \underline{55.90} & \textbf{58.95} & 42.53 & -- & -- & -- & 54.74 & 29.04 \\

\cmidrule(lr){2-13}

& \multirow{5}{*}{3G}
& CRPS &0.0093 & 0.0101 & 0.0095 & \underline{0.0089} & 0.0166 & 0.0661 & 0.0661 & 0.0646 & \textbf{0.0077} & 0.0288 \\
& & ND &\underline{0.1178} & 0.1316 & 0.1287 & 0.1219 & 0.2133 & 0.6802 & 0.6791 & 0.6654 & \textbf{0.1012} & 0.3200 \\
& & NRMSE &\underline{0.1800} & 0.1877 & 0.1909 & 0.1860 & 0.2679 & 0.7025 & 0.7013 & 0.6867 & \textbf{0.1658} & 0.3805 \\
& & MAPE &\underline{12.76} & 14.65 & 14.22 & 13.32 & 24.17 & 65.99 & 65.88 & 64.74 & \textbf{10.95} & 38.29 \\
& & PICP &52.18 & 51.88 & 55.30 & \textbf{58.31} & 41.67 & -- & -- & -- & \underline{56.81} & 30.09 \\

\cmidrule(lr){2-13}

& \multirow{5}{*}{4G}
& CRPS &0.0090 & 0.0097 & 0.0092 & \underline{0.0086} & 0.0162 & 0.0670 & 0.0669 & 0.0655 & \textbf{0.0074} & 0.0274 \\
& & ND &\underline{0.1160} & 0.1298 & 0.1268 & 0.1202 & 0.2113 & 0.6923 & 0.6911 & 0.6775 & \textbf{0.0994} & 0.2959 \\
& & NRMSE &\underline{0.1783} & 0.1865 & 0.1893 & 0.1847 & 0.2662 & 0.7130 & 0.7117 & 0.6974 & \textbf{0.1638} & 0.3536 \\
& & MAPE &\underline{12.59} & 14.45 & 14.04 & 13.16 & 23.90 & 67.33 & 67.21 & 66.03 & \textbf{10.78} & 32.58 \\
& & PICP &51.98 & 51.83 & 55.30 & \textbf{58.49} & 42.01 & -- & -- & -- & \underline{56.50} & 28.18 \\

\cmidrule(lr){2-13}

& \multirow{5}{*}{5G}
& CRPS &0.0085 & 0.0091 & 0.0086 & \underline{0.0081} & 0.0149 & 0.0505 & 0.0505 & 0.0494 & \textbf{0.0071} & 0.0241 \\
& & ND &\underline{0.1246} & 0.1387 & 0.1360 & 0.1296 & 0.2231 & 0.6435 & 0.6435 & 0.6315 & \textbf{0.1119} & 0.3413 \\
& & NRMSE &\underline{0.1923} & 0.1995 & 0.2044 & 0.1989 & 0.2816 & 0.6692 & 0.6692 & 0.6562 & \textbf{0.1823} & 0.3899 \\
& & MAPE &\underline{13.24} & 15.17 & 14.76 & 13.93 & 24.25 & 62.11 & 62.11 & 61.15 & \textbf{11.93} & 39.23 \\
& & PICP &53.82 & 52.61 & 55.88 & \textbf{57.67} & 40.72 & -- & -- & -- & \underline{57.38} & 25.59 \\

\bottomrule
\end{tabular}
}
\end{table*}

\section{Numerical Results and Discussions} \label{sec:results}
This section presents a comprehensive evaluation of the proposed EMFusion model against state-of-the-art benchmarks. To assess forecasting performance, we define both deterministic (point) and probabilistic forecasts metrics followed by the key baselines and evaluation settings.

\subsection{Evaluation Metrics}
\label{sec:evaluation_metrics}
Let \( y_i \) denote the true observed value at time step \( i \), 
\( \hat{y}_i \) the corresponding median (p50) point forecast produced by the model, 
and \( m \) the total number of samples in the test set. In what follows, we describe the point forecast and

\subsubsection{Point Forecast Metrics} evaluate the accuracy of a single predicted value (the median) against the true value. The considered metrics are listed as follows:

$\bullet$ \textbf{Mean Absolute Percentage Error (MAPE):}  measures the average percentage  difference between the predicted and actual values, and is defined as follows:
    \begin{equation}
        \text{MAPE} = \frac{100}{m} \sum_{i=1}^{m} \frac{|y_{i} - \hat{y}_{i}|}{|y_{i}|}
    \end{equation}

$\bullet$  \textbf{Normalized Deviation (ND):} 
ND metric measures the total absolute error relative to the total magnitude of the true values,
is defined as follows:
    \begin{equation}
        \text{ND} = \frac{\sum_{i=1}^{m} |y_{i} - \hat{y}_{i}|}{\sum_{i=1}^{m} |y_{i}|}
    \end{equation}
$\bullet$  \textbf{Root Mean Square Error (RMSE) and Normalized RMSE (NRMSE):} RMSE measures the standard deviation of the prediction errors. NRMSE normalizes RMSE by the root mean square of the true values, facilitating comparison across datasets with different scales.
    \begin{equation}
        \text{RMSE} = \sqrt{\frac{1}{m} \sum_{i=1}^{m} (y_{i} - \hat{y}_{i})^2}
    \end{equation}
    \begin{equation}
        \text{NRMSE} = \frac{\text{RMSE}}{\sqrt{\frac{1}{m} \sum_{i=1}^{m} y_{i}^2}}
    \end{equation}

\subsubsection{Probabilistic Forecast Metrics}
Probabilistic metrics evaluate the quality of the predicted \textit{intervals} and the \textit{full distribution}, assessing both sharpness (the width of the interval) and reliability (the coverage). 

$\bullet$ \textbf{Prediction Interval Coverage Probability (PICP):} 
measures the proportion of realized values that fall within the predicted interval. For each forecast horizon step $k$, let the prediction interval be
$
PI_\gamma = [L_k, U_k],
$
where
$
\underline{\alpha} = \frac{1-\gamma}{2},
\qquad
\overline{\alpha} = 1 - \underline{\alpha}.
$
For example, an 80\% prediction interval corresponds to $\gamma=0.8$,
yielding $\underline{\alpha}=0.1$ and $\overline{\alpha}=0.9$.
{
Ideally, $PI_\gamma$ should closely match the target
confidence level $\gamma$.}
Define the coverage indicator as follows:
\begin{equation}
c_i =
\begin{cases}
1, & \text{if } y_i \in [L_i, U_i], \\[4pt]
0, & \text{otherwise}.
\end{cases}
\end{equation}
Over {$m$ forecasting values}, the PICP is computed as
\begin{equation}
\mathrm{PICP}
=
\frac{1}{m}
\sum_{i=1}^{m} c_i.
\end{equation}
For reporting in percentage form, $100 \times \mathrm{PICP}$ is used.
Ideally, $\mathrm{PICP} \approx \gamma$, indicating {uncertainty-aware}
predictive intervals.

$\bullet$ \textbf{Continuous Ranked Probability Score (CRPS):} evaluates the quality of the 
entire predictive distribution by measuring the discrepancy between the predictive 
CDF \(F\) and the observed value \(y\) as shown below:
\begin{equation}
        \mathrm{CRPS}(F,y)
    = \int_{-\infty}^{\infty} \bigl(F(z) - \mathbf{1}\{y \le z\}\bigr)^2 \, \mathrm{d}z,
\end{equation}

where \(\mathbf{1}\{\cdot\}\) denotes the indicator function. When only predictive 
samples \(X^{(1)},\ldots,X^{(S)} \sim F\) are available, we approximate \(F\) using the  empirical CDF, i.e.,
\begin{equation}
        \widehat{F}(z) = \frac{1}{S} \sum_{s=1}^S \mathbf{1}\{X^{(s)} \le z\}
\end{equation}

and compute CRPS at each time step accordingly \cite{RasulEtAl2021TimeGrad}.

\subsection{Considered Baselines}
\label{sec:baselines}

{
To validate the performance of EMFusion, we conduct a comparative analysis against a suite of forecasting models. A brief description\footnote{{ The considered baselines  are inherently unconditional and do not incorporate external contextual information. For fair comparison, we consider unconditional version of EMFusion (\textbf{Table III}) under identical contextual settings. All EMFusion variants in \textbf{Tables IV and V} share the same backbone architecture, parameter count, optimization settings, and training protocol.}
} of each basline model is provided below:
\begin{itemize}

    \item \textbf{DDPM~\cite{ref:diffusion-energy}:} 
    the model uses a U-Net denoiser with time-dependent noise embeddings and is trained to match the true forward diffusion process via  noise-prediction. 

    \item \textbf{WGAN~\cite{huang2025probabilistic}:} 
    the generator maps historical EMF trajectories and latent noise to future EMF scenarios, while the critic   enforces distributional realism through the Wasserstein loss with gradient penalty. 

    \item \textbf{VAE~\cite{ref:vae}:} 
    the encoder maps historical EMF sequences into a low-dimensional latent space, and the decoder reconstructs future EMF trajectories from sampled latent variables conditioned on the history. This yields a parametric predictive distribution via the latent Gaussian prior.

    \item \textbf{Normalizing Flow (NF) \cite{ref:diffusion-energy}:} 
   Starting from a base distribution (e.g., multivariate Gaussian), a sequence of invertible transformations is learned to model the joint distribution of future EMF values given the past values, enabling exact likelihood evaluation and sampling.

    \item \textbf{Improved Quantile LSTM (IQLSTM):}  follows \cite{ref:iqlstm}, where an LSTM-based temporal encoder outputs multiple conditional quantiles of the future EMF distribution. The model is trained using a pinball (quantile) loss over a set of pre-defined quantile levels, providing a non-parametric representation of forecast uncertainty for each time step.

    \item \textbf{Dropout-based Probabilistic Model:} Following the Monte Carlo Dropout approach of Gal and Ghahramani \cite{ref:dropout}, this algorithm employs a deep LSTM forecaster with dropout layers activated both during training and inference, byy running multiple stochastic forward passes with different dropout masks

    \item \textbf{TimeGrad:} TimeGrad \cite{RasulEtAl2021TimeGrad} is a diffusion-based time-series forecasting model that combines an autoregressive RNN backbone with a conditional diffusion head. 
\end{itemize}
}

\subsection{Evaluation Settings}
All models were trained using an identical training dataset. Their respective hyperparameters were optimized via a grid search procedure on the validation datasets. All algorithms were executed using PyTorch on the Trillium high-performance computing cluster, hosted by SciNet at the University of Toronto. Specifically, we utilized the GPU-accelerated nodes, which are equipped with an AMD EPYC 9654 (Zen 4) 2.4 GHz CPU, 749 GB of system RAM, and four NVIDIA H100 SXM GPUs (80 GB memory each).

{ In our implementation, each 24-hour period of the EMF time series consists of $192$ regularly sampled observations. The model utilizes a 7-day lookback window to forecast the EMF trajectory for the subsequent day. Formally, this configuration yields a historical input window of $H = 1344$ time steps and a forecast horizon of $F = 192$ steps. The dataset is organized into sequential chunks, where each chunk comprises a 7-day lookback window ($H=1344$) and a 1-day forecast horizon ($F=192$). We allocate 60\% of these chunks for training, with 20\% reserved for validation and testing, respectively.  The PICP confidence level is $\gamma=80\%$.

To construct the probabilistic prediction intervals during the inference, we generate $\hat{N}=100$ independent scenarios. The structural hyperparameters for {EMFusion} are optimized to capture complex temporal dependencies. Specifically, the diffusion process is configured with $T=200$ steps by default, while the U-Net backbone employs a hierarchical depth across its layers, {with the dimension expansion factors $[1, 2, 2, 2, 2, 2, 2, 2]$ respectively.} For training, we employ a learning rate of $5 \times 10^{-4}$ and a batch size of 64 over $1500$ epochs. The architecture further incorporates an 8-head attention mechanism within a U-Net model of depth 6 to ensure adaptive feature extraction. Finally, the simulation incorporates essential physical constants, including the free-space wave impedance $Z_0 = 376.73\ \Omega$ and the antenna-gain calibration constant $A_{GC} = 9.73$.
The training procedure incorporates standard regularization  (e.g., early stopping based on validation loss, $L_2 $ regularization, and batch normalization).
}

\subsection{Performance Evaluation}

\subsubsection{Comparative Analysis of EMFusion without External Conditions}
\textbf{Table-\ref{tab:uv_mv_common_metrics_5m}} analyzes the univariate forecasting performance of the  EMFusion model, configured with imputation and without external conditions. We compare it against several baselines, including generative models (NF, DDPM, VAE, WGAN) and other probabilistic forecasting models (IQLSTM, LSTM-Dropout). The evaluation is based on five key metrics, CRPS, ND, NRMSE, and MAPE (where lower is better) and PICP (where higher is better).

In both the operator-wise and technology-band-wise breakdowns, EMFusion consistently achieves the best scores for CRPS, ND, NRMSE, and MAPE, while only WGAN achieves a fractionally better NRMSE for the VF dataset. From the operator perspective, the performance gap is even more pronounced. EMFusion's CRPS (0.0067) and MAPE (14.18\%) are far superior to all baselines in terms of TIM.
We also observe EMFusion drastically reduces all error metrics in the 2G 3G and 4G dataset. While 5G dataset appears to be the most challenging, with higher error values for all models. Despite this, EMFusion still delivers the best performance in CRPS (0.0019), ND (0.1518), and MAPE (11.57\%). 

In terms of uncertainty estimation, the highest PICP scores are consistently achieved by DDPM and NF. For example, DDPM achieves 90.22\% for VF and 82.02\% for 2G, while NF scores 72.44\% for W3. These results indicate that a larger proportion of the observed values falls within the corresponding prediction intervals. EMFusion achieves PICP values generally in the 50--60\% range, indicating under-coverage relative to the nominal 80\% confidence level. Consequently, while EMFusion improves forecasting accuracy and probabilistic forecasting metrics such as CRPS, its prediction intervals should be interpreted as empirical uncertainty estimates rather than fully calibrated uncertainty intervals.

Fig.~\ref{fig:8in1_baseliness} highlights the performance of multivariate EMFusion across all datasets organized by cellular technology bands and operators. Multivariate EMFusion with  working hour condition provides the closest match to the ground truth EMF values, i.e., it  reproduces the
daily exposure cycle, aligns well with the timing of peaks and troughs, and preserves the moderate intra-day variability without following high-frequency noise. 
Univariate EMFusion with  working hour condition exhibits a similar pattern but tends to slightly
underestimate the most pronounced peaks in some panels.
In contrast, NF and VAE generally produce smoother
forecasts that track the overall trend but miss sharp local excursions, leading
to significant phase and amplitude mismatches. DDPM is more
responsive to transient fluctuations, but this often manifests as spurious
oscillations and overfitting to local noise, especially in higher-exposure
bands such as 4G and 5G.

{

\begin{table}[t]
\centering
\small
\caption{\textbf{Performance comparison} Lower is better for ND, NRMSE, and MAPE; for PICP, a value closest to 80\% is preferred. Best per column in \textbf{bold}.}
\label{tab:merged_comparison}
\resizebox{\linewidth}{!}{
\setlength{\tabcolsep}{0.8em}
\begin{tabular}{l@{\hspace{3pt}}lcccc}
\toprule
\multicolumn{2}{c}{\textbf{Method}} &
\textbf{ND} &
\textbf{NRMSE} &
\textbf{MAPE} &
\textbf{PICP} \\
\midrule

\multicolumn{2}{l}{NF}      & 0.1739 {\tiny$\pm$ 0.0332} & 0.2447 {\tiny$\pm$ 0.0941} & 18.1875 {\tiny$\pm$ 2.8910} & 69.1871 {\tiny$\pm$ 6.0702} \\
\multicolumn{2}{l}{IQLSTM}  & 0.1829 {\tiny$\pm$ 0.0338} & 0.2575 {\tiny$\pm$ 0.0954} & 19.2144 {\tiny$\pm$ 2.6370} & 61.8676 {\tiny$\pm$ 5.9028} \\
\multicolumn{2}{l}{WGAN}    & 0.1959 {\tiny$\pm$ 0.0507} & 0.2588 {\tiny$\pm$ 0.0998} & 21.3099 {\tiny$\pm$ 4.3154} & 50.0930 {\tiny$\pm$ 9.8127} \\
\multicolumn{2}{l}{VAE}     & 0.1908 {\tiny$\pm$ 0.0411} & 0.2597 {\tiny$\pm$ 0.0971} & 20.6526 {\tiny$\pm$ 3.8411} & 52.2893 {\tiny$\pm$ 4.4282} \\
\multicolumn{2}{l}{DDPM}    & 0.1989 {\tiny$\pm$ 0.0520} & 0.2647 {\tiny$\pm$ 0.0985} & 22.5219 {\tiny$\pm$ 6.0977} & \textbf{85.9521} {\tiny$\pm$ 6.4352} \\
\multicolumn{2}{l}{Dropout} & 0.2418 {\tiny$\pm$ 0.0481} & 0.2937 {\tiny$\pm$ 0.0867} & 30.2881 {\tiny$\pm$ 6.8275} & 27.9310 {\tiny$\pm$ 6.7606} \\

\cmidrule(lr){1-6}

\multirow{5}{*}{\rotatebox{90}{\scriptsize\textbf{EMFusion-MV}}}
& Uncondition  & 0.2486 {\tiny$\pm$ 0.0688} & 0.3257 {\tiny$\pm$ 0.1048} & 23.6121 {\tiny$\pm$ 6.2094} & 33.0397 {\tiny$\pm$ 10.4381} \\
& Multi        & 0.2083 {\tiny$\pm$ 0.0412} & 0.2808 {\tiny$\pm$ 0.0955} & 22.8826 {\tiny$\pm$ 5.1239} & 41.6521 {\tiny$\pm$ 6.5349} \\
& WorkingDay   & 0.6345 {\tiny$\pm$ 0.2315} & 0.6705 {\tiny$\pm$ 0.2088} & 62.0675 {\tiny$\pm$ 22.9755} & 4.8629 {\tiny$\pm$ 7.7988} \\
& Season       & 0.6355 {\tiny$\pm$ 0.2347} & 0.6735 {\tiny$\pm$ 0.2123} & 61.9106 {\tiny$\pm$ 23.5813} & 5.7850 {\tiny$\pm$ 8.1614} \\
& WorkingHour  & \textbf{0.1049} {\tiny$\pm$ 0.0241} & \textbf{0.1858} {\tiny$\pm$ 0.0936} & \textbf{10.7561} {\tiny$\pm$ 1.4898} & 56.4799 {\tiny$\pm$ 5.0838} \\

\bottomrule
\end{tabular}
}
\end{table}

\subsubsection{Univariate vs Multivariate Conditional EMFusion}

\textbf{Table-\ref{tab:forecasting_comparison}} shows the significance of incorporating appropriate contextual conditions in the reverse diffusion process. In particular, the \textbf{WorkingHour}-conditioned EMFusion  achieves the lowest CRPS, ND, NRMSE, and MAPE  regardless of the operator (W3, Iliad, TIM, VF) 
or technology band (2G--5G). The multi-conditioned multivariate model (last column) does not surpass the 
\textbf{WorkingHour}-only model, suggesting that fine-grained time-of-day conditioning captures most of the useful temporal structure for EMF exposure, while additional context yields diminishing returns. From the operator perspective, TIM and VF benefit strongly from conditional diffusion: for example, under multivariate modeling their CRPS and MAPE values under EMFusion (WorkingHour) are an order of magnitude smaller than uncondition EMFusion, indicating substantially sharper and more accurate forecasts.

Table~\ref{tab:forecasting_comparison} shows a clear dominance of \texttt{WorkingHour} conditioning, while multi-conditioning  often yields diminishing or negative returns. This reflects the characteristics of the University Hospital site, where EMF exposure closely follows  visitor activity with a regular diurnal pattern, strong daytime peaks and low nighttime levels. As a result, \texttt{WorkingHour} captures the dominant temporal structure of BS power allocation and explains most of the variance in this dataset.
Contextual relevance, however, is environment-dependent. In residential areas, traffic peaks may shift to evenings and weekends, increasing the importance of \texttt{WorkingDay} or holiday-related features.
The performance degradation observed under multi-conditioning suggests that expanding the conditioning space without sufficient diversity in the training data can reduce learning efficiency. Higher-dimensional conditioning requires adequate coverage of contextual state combinations; when certain states (e.g., \texttt{WorkingDay}) are underrepresented in the dataset, the cross-attention mechanism struggles to prioritize the most informative signals.
}

From the technology-band viewpoint, EMFusion sharply reduces all error metrics for legacy bands (2G, 3G, 4G), with the largest relative gains observed for 4G, where EMFusion (\textbf{WorkingHour}) approximately halves the CRPS and MAPE compared to unconditioned EMFusion in the multivariate setting. The 5G band remains the most challenging case, with uniformly higher errors across all models due to its more volatile traffic and deployment patterns. Nonetheless, EMFusion still delivers the best multivariate performance in CRPS, ND, and MAPE for 5G, indicating that the diffusion-based approach remains adaptive even under rapidly fluctuating EMF dynamics. In terms of interval coverage, unconditioned EMFusion exhibits very low PICP values in the multivariate case, reflecting overly narrow and under-dispersed prediction intervals. EMFusion, by contrast, achieves PICP values predominantly in the 50--65\% range across operators and bands, with the \textbf{WorkingHour} configuration providing the strongest overall forecasting performance. Although these values remain below the nominal 80\% target, the resulting prediction intervals exhibit under-coverage. Therefore, the results should be interpreted as demonstrating improved probabilistic forecasting performance rather than fully calibrated uncertainty estimates. Achieving calibrated uncertainty estimates remains an important direction for future work.

\textbf{Table-\ref{tab:merged_comparison}} reports the forecasting performance of all methods, averaged over all carriers and technologies. The proposed \textbf{EMFusion-MV-WorkingHour} achieves the lowest ND, NRMSE, and MAPE among all compared methods, indicating superior forecasting accuracy. However, its PICP is lower than that of DDPM and NF and remains below the nominal 80\% target. Therefore, while EMFusion delivers the best overall forecasting accuracy, its prediction intervals exhibit under-coverage and should be interpreted as empirical uncertainty estimates rather than fully calibrated prediction intervals.

\subsubsection{Inference Complexity Analysis}
{
Given a look-back window of length $H$, a forecast horizon $F$, an embedding dimension $d$, and $T$ total diffusion steps, we analyze the inference complexity as shown below. The inference complexity is dominated by the $T$ iterative denoising steps within the U-Net backbone and the cross-attention mechanism.
For each denoising step, the complexity of the convolutional layers and attention projections scales with the total sequence length $(H+F)$, the number of variates $N$ and the number of EMF trajectories $\hat{N}$. Specifically, the complexity of a single denoising step in our multivariate architecture is $\mathcal{O}(\hat{N} \cdot N \cdot (H+F) \cdot d^2)$. Consequently, the total inference complexity for generating a multivariate forecast scales as
$
    \mathcal{T}_{\text{inference}} = \mathcal{O}(T \cdot \hat{N} \cdot N \cdot (H + F) \cdot d^2).
$
The inference time for EMFusion is in the order of milliseconds. { This supports scalable inference, though real-time deployment would additionally require operator-side integration and latency testing.}
}

\subsubsection{Ablation Studies}

\begin{table}[htbp]

\centering
\small
\caption{Ablation study on diffusion step size and U-Net depth in EMFusion. The training times reported in the tables are formatted as \textbf{HH:MM:SS}, where \textbf{HH}, \textbf{MM} and \textbf{SS} represents the number of hours, minutes, seconds, respectively. }
\resizebox{\linewidth}{!}{
\begin{tabular}{llccccc}
\toprule
\multicolumn{2}{c}{\textbf{Setting}} & ND & NRMSE & MAPE & PICP & Time\\
\midrule\midrule

\multirow{3}{*}{\rotatebox{90}{\scriptsize{\shortstack{Diffusion\\Steps}}}}
& 50  & 0.2393 & 0.3963 & 31.62 & 56.94 & \textbf{01:32:20} \\ 
& 200 & 0.1335 & 0.2660 & 15.44 & 65.29 & 02:02:41 \\
& 300 & 0.1759 & 0.3310 & 22.81 & 68.08 & 02:10:18 \\

\cmidrule(lr){1-7}

\multirow{3}{*}{\rotatebox{90}{\scriptsize{\shortstack{U-Net\\Depth}}}}
& 3 & 0.1786 & 0.3343 & 22.53 & 73.98 & \textbf{02:27:50} \\
& 7 & 0.1872 & 0.3393 & 25.92 & 72.03 & 02:51:08 \\
& 9 & \textbf{0.1625} & \textbf{0.3246} & \textbf{19.03} & 62.70 & 03:01:51 \\

\bottomrule
\end{tabular}
}
\label{tab:ablation_emfusion}
\end{table}

\begin{table}[htbp]

\centering
\small
\caption{ Rolling window evaluation for EMFusion.}
\resizebox{\linewidth}{!}{
\begin{tabular}{lcccc}
\toprule
\textbf{Period} & \textbf{ND} & \textbf{NRMSE} & \textbf{MAPE} & \textbf{PICP} \\
\midrule

2023/07/26 -- 2023/12/12 & 0.1510 & 0.2040 & 20.25 & 55.17 \\

2023/08/25 -- 2024/01/11 & 0.1741 & 0.2374 & {18.16} & 56.69 \\
2023/09/24 -- 2024/02/10 & 0.1390 & {0.1954} & 15.84 & 59.76 \\
2023/10/24 -- 2024/03/12 & 0.1959 & 0.3737 & 21.96 & 61.17 \\

\midrule
\textbf{Average} & 0.1650 & 0.2526 & 19.05 & 58.20 \\
Full Dataset (No rolling window) & {0.1335} & 0.2660 & 15.44 & {65.29} \\

\bottomrule
\end{tabular}
}
\label{tab:time_window_emfusion}
\end{table}
In \textbf{Table~\ref{tab:ablation_emfusion}},
 we varied the diffusion step size $\{50, 200, 300\}$ while fixing the U-Net depth at 8. 
Also, we investigated the impact of the U-Net backbone depth across $\{3, 7, 9\}$ layers while maintaining a fixed diffusion step size of $50$.
Overall, performance variations across 50–300 steps are relatively small, indicating that the model is not highly sensitive to the diffusion steps.  
{However, the depth of UNet architecture does impact the  performance, i.e., the model with 9 layers outperform. This is because the deeper the model, the larger its receptive field across the entire time series, enabling it to capture abstract information within long sequences at the expense of additional complexity.

{To evaluate the framework's stability across distinct temporal segments, we then performed a rolling-window evaluation by sliding 5-month window across the 8-month period. The results in Table~\ref{tab:time_window_emfusion} demonstrate consistent performance, confirming that {EMFusion} remains adaptive to the temporal variations. We note that the reduction in training data length inherent in the rolling-window setup can slightly impact performance; for instance, the average MAPE increases from 15.44\% to 19.05\% compared to the full dataset. This marginal degradation is attributed to the limited training data available within a window.}
}
\begin{figure}
    \centering
    \includegraphics[width=1\linewidth]{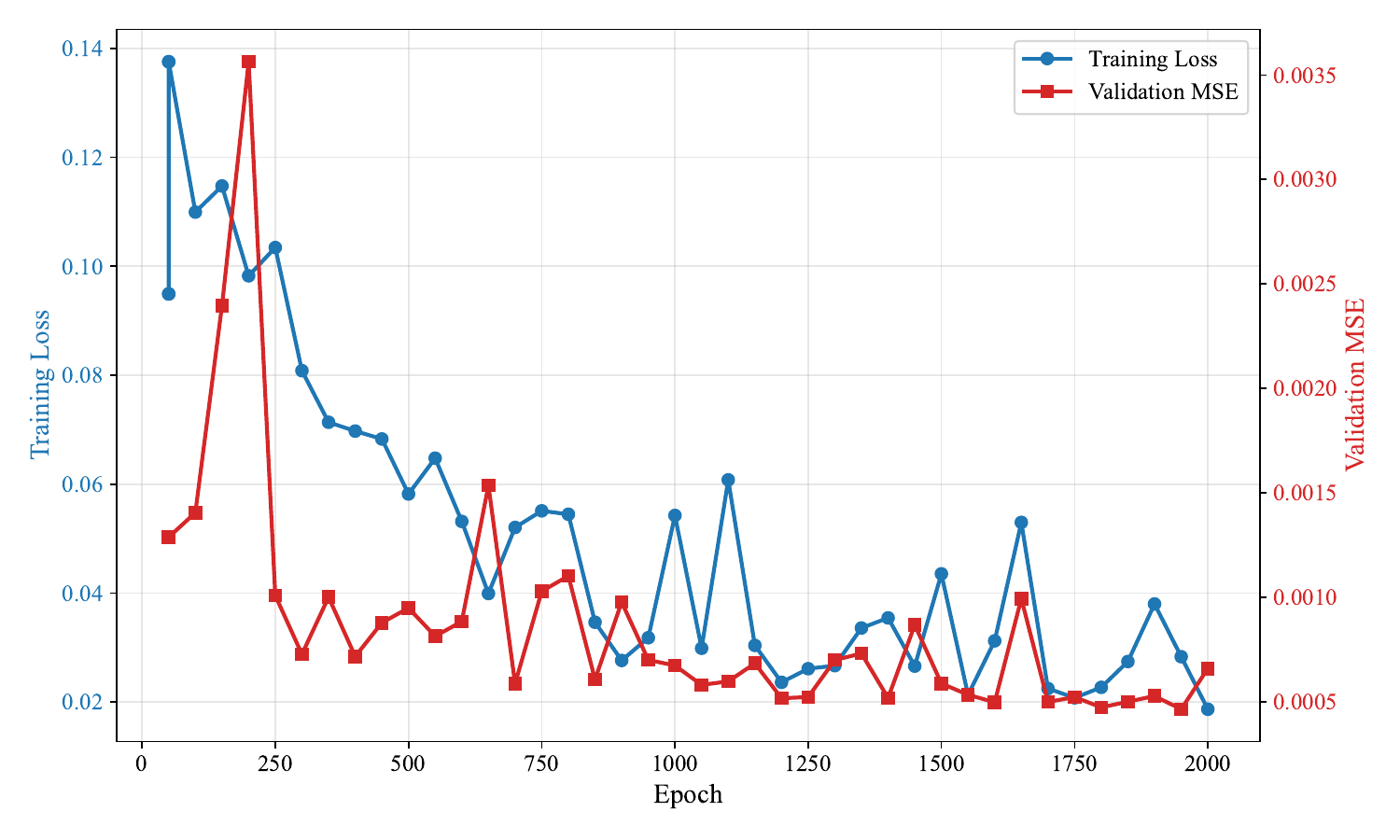}
\caption{Training and validation loss  for the W3 dataset.}
    \label{fig:training_loss_vs_validation_loss}
    \vspace{-3mm}
\end{figure}
To address the risk of potential overfitting, Fig.~\ref{fig:training_loss_vs_validation_loss} illustrates the training and validation loss trajectories. The synchronized downward trend and stable convergence of both metrics demonstrate consistent generalization to unseen data without significant divergence. 
The model also includes explicit regularization through $L_2$ weight decay and Group Normalization, which help control model capacity and stabilize training. In addition, the diffusion objective perturbs inputs with Gaussian noise at multiple levels while sharing parameters across diffusion steps, introducing an implicit regularization effect that {improves the model's tolerance to noisy input conditions}.

{
\subsection{Ethical and Deployment Considerations}

\subsubsection{Data Governance}
{Uncertainty-aware} EMF forecasting requires adaptive data governance. In this study, data acquisition and analysis were conducted by the same organization, and the raw datasets were released as open source to ensure transparency and reproducibility. In multi-agency settings, strict compliance with regulations such as the General Data Protection Regulation (GDPR) \cite{protection2018general} is essential, including the obfuscation of sensitive metadata and secure data-sharing protocols to prevent unauthorized access or misuse.

\subsubsection{Mitigating Monitoring Biases}
Long-term monitoring can suffer from sampling biases due to sensor outages or irregular measurement intervals. EMFusion addresses this through its imputation-based sampling strategy (Section \ref{subsec:imputation_sampling}), treating missing data as a structural inpainting task to preserve temporal coherence. While this reduces algorithmic bias, future monitoring systems should incorporate hardware-level detectors to identify abnormal sensor behavior.

\subsubsection{Scalability for Nationwide Deployment}
{
Nationwide deployment requires balancing computational overhead with monitoring density. The inference complexity of EMFusion is given by $\mathcal{O}(\hat{N} \cdot N \cdot (H+F) \cdot d^2)$, indicating that the computational cost scales linearly with respect to the forecast horizon $F$. However, practical nationwide implementation also requires supporting infrastructure for real-time data ingestion, distributed processing, and long-term cloud storage.
}
}
\section{Conclusion} 
\label{sec:Conclusion}
This paper introduced EMFusion, a conditional diffusion-based probabilistic forecasting framework for multivariate narrow-band EMF exposure forecasting. By leveraging conditional embeddings, cross-attention-based context integration, and imputation-based sampling, EMFusion improves forecasting accuracy while providing uncertainty quantification. Experiments on a real-world EMF dataset demonstrate that EMFusion outperforms representative deep learning and generative forecasting baselines in both deterministic and probabilistic metrics.
EMFusion is a general data-driven framework that can be adapted to other locations when monitoring data and contextual information are available. However, the current evaluation is limited to a single-site measurement campaign at the University Hospital Tor Vergata in Rome, and the reported rolling-window analysis demonstrates temporal consistency within this environment rather than cross-site generalization. Consequently, the present results should be interpreted as validation on a single urban monitoring site. Future work will investigate diverse multi-site datasets, richer contextual features, broader assessments of model generalization across deployment environments, and the integration of EMFusion predictions into EMF-aware optimization and control applications, together with closed-loop network-control validation.

{
\section*{Acknowledgment}
The authors would like to thank Dr. Daniele Franci and Settimio Pavoncello for providing the EMF measurement hardware
used to collect the Italian dataset.
}
\bibliographystyle{IEEEtran}
\bibliography{bibi-clean2}
\vspace{-0.1in}
\begin{IEEEbiography}[{\includegraphics[width=0.9in,height=1.1in,clip,keepaspectratio]{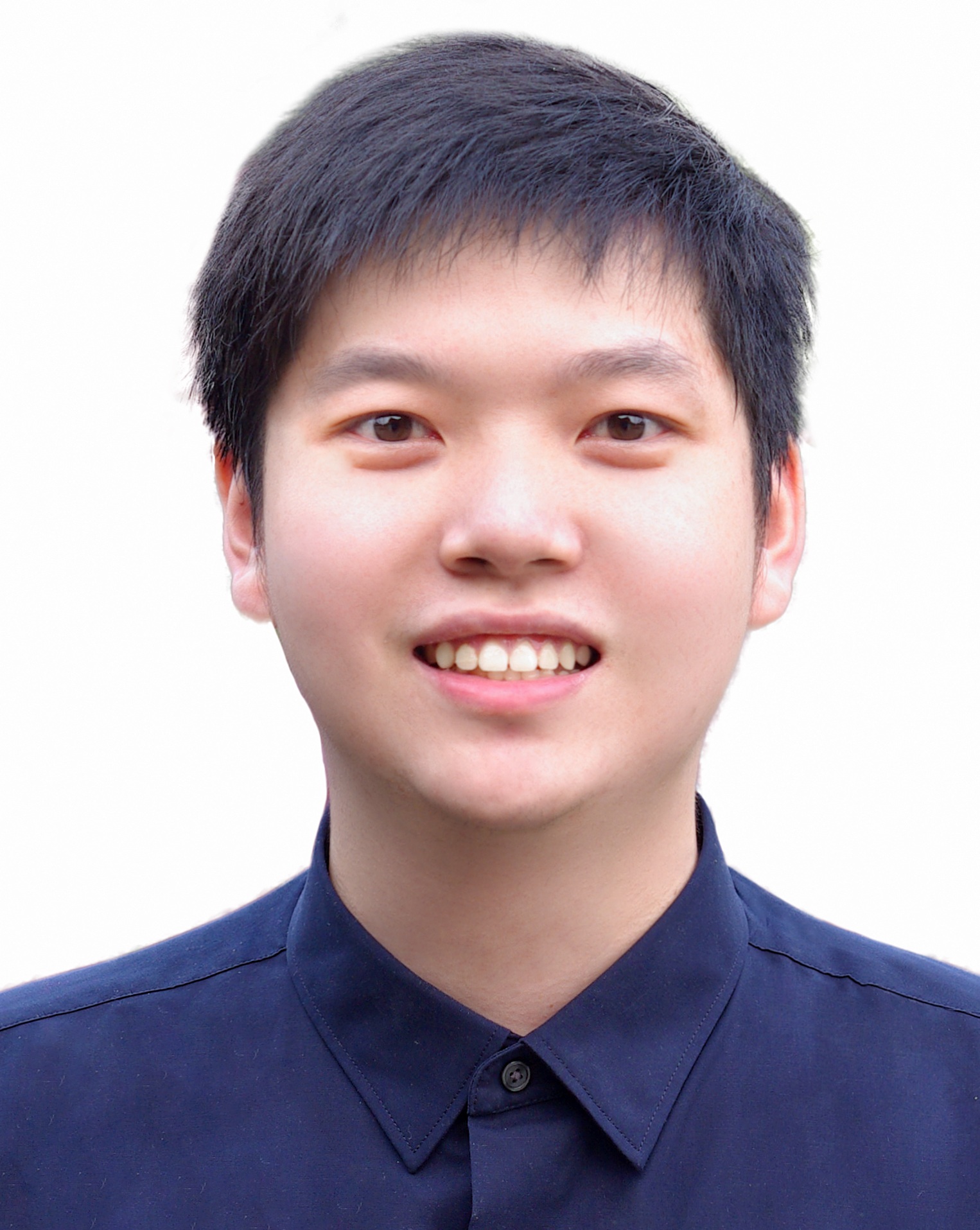}}]{Zijiang Yan} 
 ({Graduate Student Member}, IEEE) received the B.S. degree with a double major in Computer Science and Statistics from York University, Toronto, ON, Canada, in 2021. He is currently pursuing the M.A.Sc. degree in Electrical Engineering and Computer Science at York University. His research interests include AI-enabled communications, quantum machine learning, diffusion models, large language models, and intelligent wireless networking. He received third place in the 2025 Student Innovation Competition on Sustainable Space Communications, organized by the Satellite and Space Communications Committee (SSC) of the IEEE Communications Society (\textsc{IEEE ComSoc}). He was also the recipient of the Lassonde Undergraduate Research Award (LURA) from York University in 2021.
\end{IEEEbiography}
\vspace{-0.1in}
\begin{IEEEbiography}[{\includegraphics[width=0.9in,height=1.1in,clip,keepaspectratio]{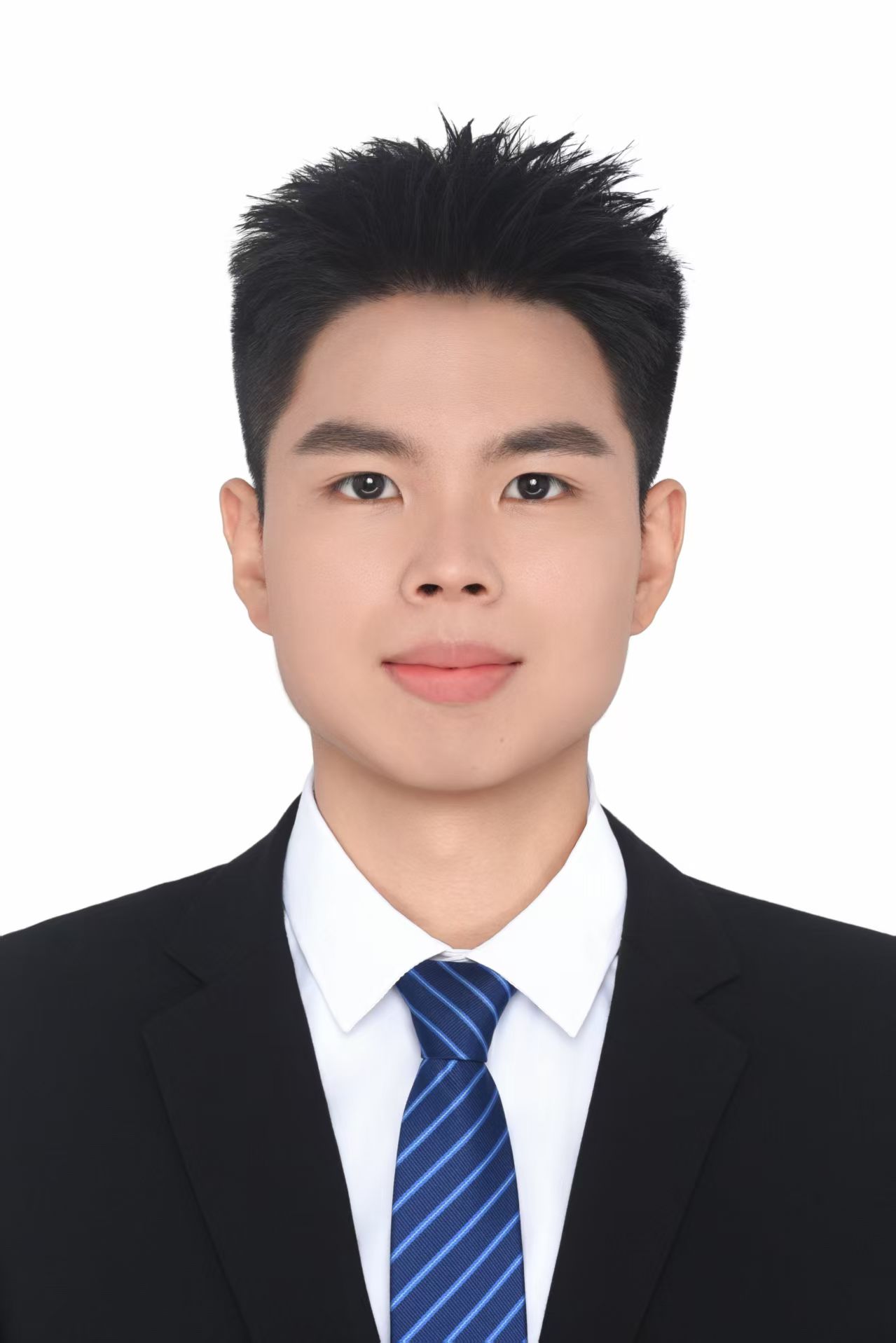}}]{Yixiang Huang} \,
 ({Graduate Student Member}, IEEE) 
 received the B.Eng. and M.Eng. degrees in electrical engineering from Huazhong University of Science and Technology (HUST), Wuhan, China, in 2023 and 2026, respectively. He is currently with China Southern Grid Electric Power Dispatch Control Center. His research interests include power system analysis, probabilistic load forecasting, and artificial intelligence applications for time series.
\end{IEEEbiography}
\vspace{-0.1in}
\begin{IEEEbiography}[{\includegraphics[width=0.9in,height=1.1in,clip,keepaspectratio]{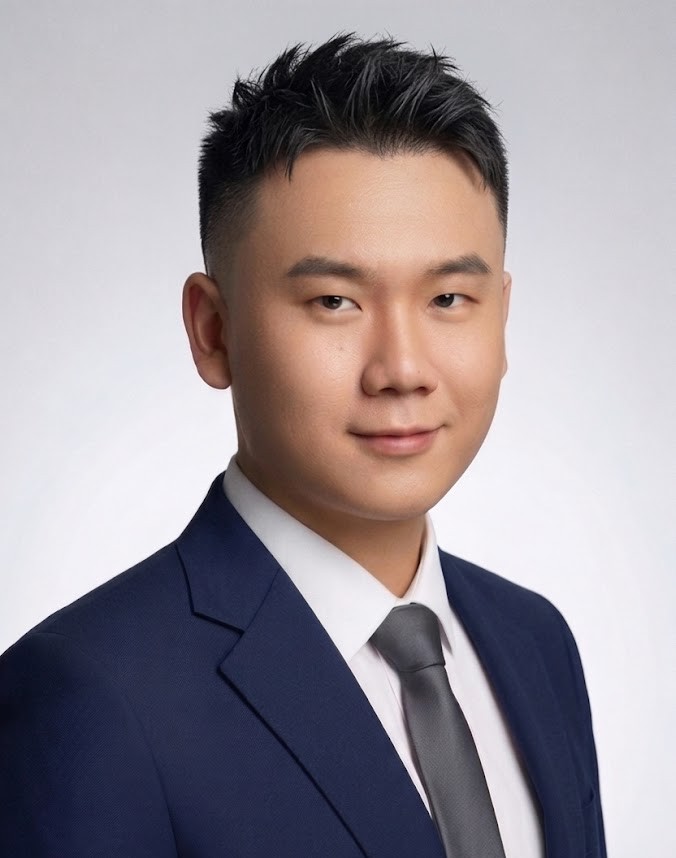}}]{Jianhua Pei} ({Member}, IEEE) received the B.Eng. and Ph.D. degrees in electrical engineering from Huazhong University of Science and Technology (HUST), Wuhan, China, in 2019 and 2025, respectively. He is currently with the Central China Branch of State Grid Corporation of China. He was also a visiting Ph.D. student at the Department of Electrical Engineering and Computer Science, Lassonde School of Engineering, York University, Canada, in 2024. His research interests include power system data quality improvement, power system dynamics, power system cybersecurity, and artificial intelligence applications for communications.
\end{IEEEbiography}

\begin{IEEEbiography}[{\includegraphics[width=0.9in,height=1.1in,clip,keepaspectratio]{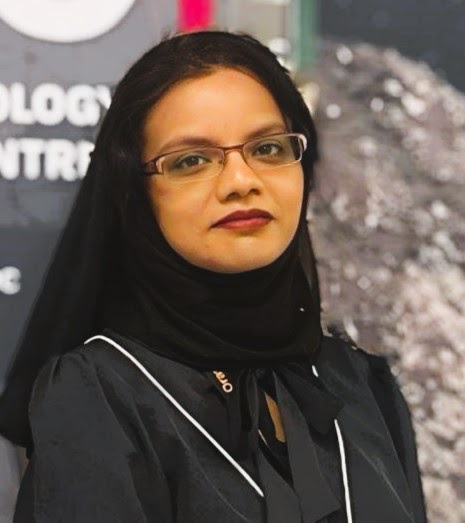}}]{Hina Tabassum}
 ({Senior Member}, IEEE) 
 received the Ph.D. degree from the King Abdullah University of Science and Technology. She is currently an Associate Professor with the Lassonde School of Engineering at York University, Canada, where she joined in 2018. She also serves as a Visiting Faculty  at University of Toronto and holds the York Research Chair in 5G/6G-enabled mobility and sensing applications. Her research focuses on the modeling, analysis, and optimization of next-generation wireless communication, localization, and sensing networks.  She has co-authored 2 books, 8 book chapters, and over 120 publications, including 88 journal articles and 42 conference papers in leading IEEE venues. She has delivered more than 20 invited tutorials and talks at international venues on wireless communications and sensing networks. She was selected as an IEEE Communications Society Distinguished Lecturer for 2025–2026 and has been listed among Stanford’s World’s Top 2\% Researchers from 2021 to 2025.   Her recognitions include the N2Women Star in Networking and Communications (2025), the N2Women Rising Star Award (2022), and the Early Career Lassonde Innovation Award (2023). She currently serves as an editor for several leading IEEE journals, including IEEE Transactions on Communications, IEEE Transactions on Mobile Computing, IEEE Transactions on Wireless Communications, and IEEE Communications Surveys and Tutorials.
\end{IEEEbiography}

\begin{IEEEbiography}[{\includegraphics[width=1in,height=1.25in,clip,keepaspectratio]{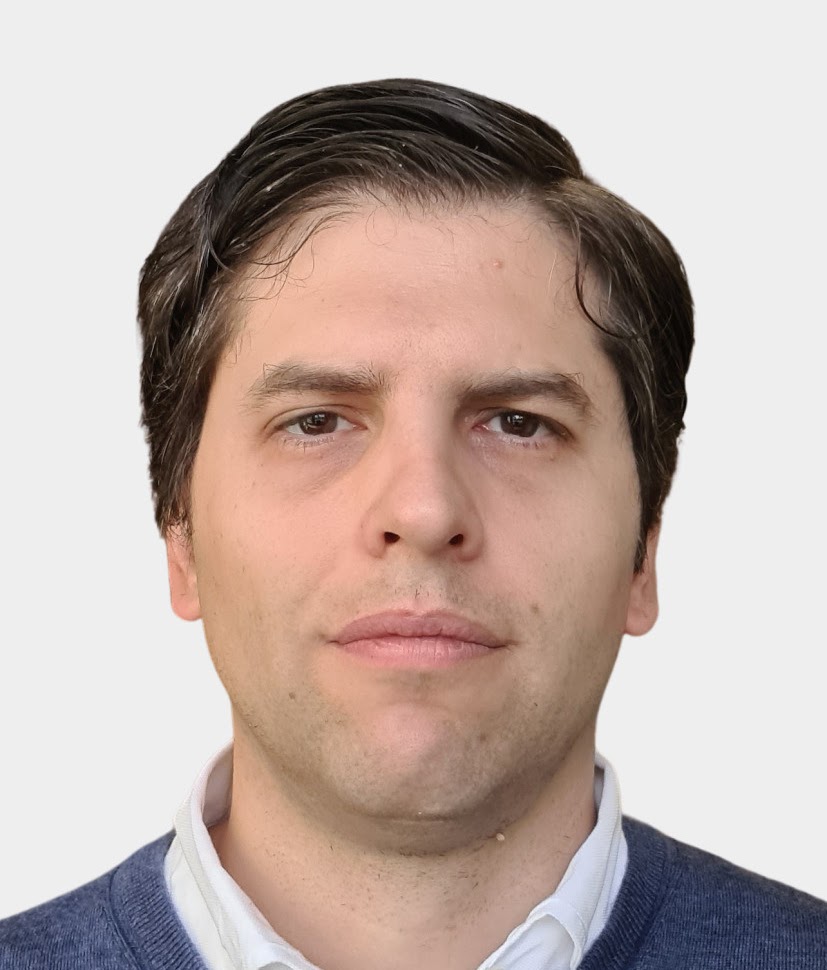}}]{Luca Chiaraviglio} 
({Senior Member}, IEEE) received the Ph.D. degree in Telecommunication and Electronics Engineering from the Polytechnic University of Turin, Italy. He is Full Professor with the University of Rome Tor Vergata, Italy. He has co-authored more than 190 papers published in international journals, books, and national/international conferences. He is the Principal Investigator of the RESTART Program, University of Rome Tor Vergata, managing a total budget of more than 10.2 MEUR. His current research interests include 5G and B5G networks, optimization applied to telecommunication networks, electromagnetic fields, and health risks assessment of 5G and B5G communications. Some of his papers are listed as the Best Readings on Green Communications by the IEEE. He has been recognized as an author in the top 2\% of world scientists according to the 2021–2025 updates of the science-wide author databases of standardized citation indicators. He has received the Best Paper Award at the IEEE Vehicular Technology Conference (VTC)-Spring 2020, the IEEE VTC-Spring 2016, and the 2018 Conference on Innovation in Clouds, Internet and Networks (ICIN), all of them appearing as the first author.
\end{IEEEbiography}

\end{document}